\documentclass[journal]{IEEEtran}
\ifCLASSINFOpdf
\else
\fi
\usepackage[utf8]{inputenc}

\usepackage{hyperref}

\usepackage[setcolon]{kmath}
\usepackage{amssymb,amsthm,amsmath}
\usepackage{mathtools,mathrsfs}
\usepackage{stmaryrd}
\usepackage{textcomp}
\usepackage{cleveref}
\usepackage{algorithmic}
\usepackage[algoruled, linesnumbered, onelanguage]{algorithm2e}
\usepackage[sort&compress,numbers]{natbib}
\usepackage{comment}
\usepackage{xfrac}    

\usepackage{xcolor}
\definecolor{rreport_blueMath}{RGB}{18,75,126}
\definecolor{rreport_cite}{RGB}{189,0,38} 
\definecolor{rreport_equa}{RGB}{8,81,156} 
\definecolor{rreport_commentary}{RGB}{35,139,69}

\definecolor{myblue}{RGB}{18,75,126}
\definecolor{CEcolor}{RGB}{0,109,44} 
\definecolor{complexity}{RGB}{153,52,4} 

\definecolor{myviolet}{RGB}{84,39,143}

\hypersetup{
  colorlinks = true,
  linkcolor  = orange!85!black,
  citecolor  = myblue!100!black,
  urlcolor   = myviolet,
  pdfauthor={},
  pdftitle={},
  pdfsubject={},
  pdfkeywords={},
  pdfproducer={pdflatex},
  pdfcreator={}
}

\usepackage{subcaption}

\SetCommentSty{mycommfont}

\usepackage{pgffor}
\foreach \x in {a,...,z}{
  \expandafter\xdef\csname bf\x \endcsname{\noexpand\ensuremath{\noexpand\mathbf{\x}}}
  \expandafter\xdef\csname bs\x \endcsname{\noexpand\ensuremath{\noexpand\boldsymbol{\x}}}
}

\foreach \x in {A,...,Z}{
  \expandafter\xdef\csname bs\x \endcsname{\noexpand\ensuremath{\noexpand\boldsymbol{\x}}}
  \expandafter\xdef\csname bf\x \endcsname{\noexpand\ensuremath{\noexpand\mathbf{\x}}}
  \expandafter\xdef\csname bb\x \endcsname{\noexpand\ensuremath{\noexpand\mathbb{\x}}}
  \expandafter\xdef\csname ds\x \endcsname{\noexpand\ensuremath{\noexpand\mathds{\x}}}
  \expandafter\xdef\csname cal\x \endcsname{\noexpand\ensuremath{\noexpand\mathcal{\x}}}
}

\newcommand{\algoFitra}{FITRA}

\newcommand{\algoPGsqueezing}{PGs}
\newcommand{\algoFW}{FW}
\newcommand{\algoFWsqueezing}{FWs}

\DeclareMathOperator{\sign}{sign}
\DeclareMathOperator{\card}{card}
\DeclareMathOperator{\krank}{kruskal}

\newcommand{\intervint}[2]{\kbrace{#1\dots#2}}

\newcommand{\coefv}{\bfx}
\newcommand{\coeffv}{\coefv}
\newcommand{\vcoeff}{\coefv}
\newcommand{\sat}{w}

\newcommand{\sateq}{\widetilde{w}}
\newcommand{\element}[1]{\kparen{#1}}

\newcommand{\cst}{\alpha}

\newcommand{\vobs}{\bfy}
\newcommand{\dimobs}{m}

\newkfunc{\lagrangian}{\mathscr{L}}

\newkfunc{\primal}{P}
\newkfunc{\primaleq}{\widetilde{P}}
\newkfunc{\dual}{D}
\newkfunc{\dualeq}{\widetilde{D}}
\newkfunc{\gap}{\mathrm{gap}}
\newkfunc{\dualscaling}{\mathrm{dualscal}}

\newcommand{\dualset}{\calU}
\newcommand{\jointdualset}{\calD}
\newcommand{\vdual}{\bfu}

\newcommand{\vdualsign}[1][\pm]{\bfv_{#1}}

\newcommand{\mOp}{\bfA}
\newcommand{\mOpv}{\bfa}

\newcommand{\idxsat}{i}

\newcommand{\xsol}{\coeffv^\star}
\newcommand{\dimx}{n}

\newcommand{\vdualsol}{\vdual^\star}

\newcommand{\satsol}{\sat^\star}             
\newcommand{\satset}{\calI}          
\newcommand{\cset}[1]{\bar{#1}}
\newcommand{\satsetsol}{\satset^\star}   
\newcommand{\sqatoms}{\bfs}            
\newcommand{\xr}{\bfq}             
\newcommand{\xrsol}{\xr^\star}         
\newcommand{\dimxr}{q}             

\newcommand{\nsatc}{p}             
\newcommand{\mOps}{\bfB}           
\newcommand{\mOpsv}{\bfb}          
\newcommand{\sreg}{\calS}            
\newcommand{\scen}{\bfc}           
\newcommand{\srad}{r}            
\newcommand{\scalf}{\alpha}          
\newcommand{\gstep}{\eta}           
\newcommand{\cstep}{\gamma}      
\newcommand{\Hessian}{\mathbf{H}}       
\newcommand{\ubsat}{\overline{\sat}}         
\newcommand{\gradf}{\mathbf{g}}        

\newcommand{\onethird}{\text{\sfrac{$1$}{$3$}}}
\newcommand{\twothird}{\text{\sfrac{$2$}{$3$}}}

\newtheorem{theorem}{Theorem}
\newtheorem{lemma}{Lemma}

\crefname{algocf}{alg.}{algs.}
\Crefname{algocf}{Algorithm}{Algorithms}

\begin{document}

%
\title{Safe Squeezing for Antisparse Coding}
%
%
%

\author{Clément Elvira and 
		  Cédric Herzet
\thanks{C. Elvira and C. Herzet are with Univ Rennes, Inria, CNRS, IRISA F-35000 Rennes, France.}
\thanks{e-mail: \texttt{prenom.nom@inria.fr}.}
\thanks{Part of this work has been funded thanks to the Becose ANR project no. ANR-15-CE23-0021 and the Labex CominLabs.}
\thanks{The research presented in this paper is reproducible. Code and data are available at \url{https://gitlab.inria.fr/celvira/safe-squeezing}.}
\thanks{}
}

%
%

\markboth{}%
{}
%



\maketitle

\begin{abstract}
   Spreading the information over all coefficients of a representation is a desirable property in many applications such as digital communication or machine learning. 
   This so-called antisparse representation can be obtained by solving a convex program involving an $\ell_\infty$-norm penalty combined with a quadratic discrepancy.
   In this paper, we propose a new methodology, dubbed safe squeezing, to accelerate the computation of antisparse representation.
   We describe a test that allows to detect saturated entries in the solution of the optimization problem.
   The contribution of these entries is compacted into a single vector, thus operating a form of dimensionality reduction.
   We propose two algorithms to solve the resulting lower dimensional problem.
   Numerical experiments show the effectiveness of the proposed method to detect the saturated components of the solution and illustrates the induced computational gains  in the resolution of the antisparse problem.
\end{abstract}

\begin{IEEEkeywords}
   antisparse coding, safe screening, scaled projected-gradient algorithm, Frank-Wolfe algorithm.
\end{IEEEkeywords}

%
\IEEEpeerreviewmaketitle


\section{Introduction}

\IEEEPARstart{I}{n} the last decades, convex optimization has become a central element in the resolution of inverse problems. 
This success revolves around  two main ingredients. 
First, it has been shown that  proper ``regularizing'' functions enforce desirable properties on the solutions of convex  problems.   
One of the first (and maybe most striking) example of such a behavior is the use of  ``$\ell_1$-norm'' regularization, promoting sparsity of the solutions.
The second ingredient which sparked the success of convex optimization is the advent of numerical procedures able to solve efficiently (up to some accuracy) problems involving thousands to billions of variables. 
To name a few, let us mention the Augmented Lagrangian methods~\cite{Rockafellar1973}, the forward-backward splitting~\cite{bruck1975}, or the Alternating direction method of multipliers~\cite{Gabay1976ADA}.
 
Of particular interest in this paper is an acceleration method first proposed by El Ghaoui \textit{et al.} in \cite{ghaoui2010safe} in the context of sparsity-promoting convex problems, namely ``\textit{safe screening}''. 
This procedure leverages two main elements.
First, the solutions of convex problems involving $\ell_1$ regularization are typically sparse, \textit{i.e.}, contains a large number of zeros.  
 For example, the solutions of the well-known ``Lasso'' problem 
\begin{equation}\label{eq:Lasso}
  \xsol \in \kargmin_{\coeffv\in\kR^n}
  \tfrac{1}{2}\kvvbar{\vobs - \mOp\coeffv}_2^2 + \lambda \kvvbar{\coeffv}_1, 
\end{equation}
where $\vobs\in\kR^m$ is an observation vector,  $\mOp\in\kR^{m\times n}$ a representation matrix and $\lambda$ a penalization parameter, are known to contain at most $m$ nonzero components, see \cite[Theorem~3.1]{Foucart2013} or \cite[Theorem~3]{rosset2004boosting}.
Second, if the position of (some of) the zeros in $\coeffv_\lambda^\star$ are known, \eqref{eq:Lasso} can be transformed into a problem of smaller dimension where the columns of $\mOp$ associated to the position of known zeros are simply discarded. Solving this reduced problem may then potentially result in  huge memory and computational savings.
As far as sparsity-promoting problems are concerned, the idea of safe screening thus consists in designing simple tests allowing to detect zeros of the solution.

Since the seminal work by El Ghaoui \textit{et al.}, safe screening has been developed and improved in many contributions of the literature, see \textit{e.g.},~\cite{fercoq2015,bonnefoy2015,malti2016,herzet2019}.
Recently, this strategy has also been extended to other families of regularizers~\cite{Ndiaye2015,Ndiaye2017,Rakotomamonjy_Gasso_Salmon19}.
However, to the best of our knowledge, all these contributions have focused so far on the resolution of ``sparsity-promoting'' problems, where the solutions contain many zeros. 
In this paper, we show that the principles ruling safe screening may be extended to another family of convex optimization problems. 
More specifically, we address the following optimization problem
 \begin{equation}
  \label{eq:antisparse_coding}
  \stepcounter{equation}
  \tag{$\theequation-\calP_\lambda^{\infty}$}
  \xsol \in \kargmin_{\coeffv\in\kR^n}
  \tfrac{1}{2}\kvvbar{\vobs - \mOp\coeffv}_2^2 + \lambda \kvvbar{\coeffv}_\infty,
\end{equation}
where the admissible range of the coefficients $\coeffv$ is penalized through an $\ell_\infty$-norm.
Here, the $\ell_\infty$-norm is advocated for spreading the information over all representation coefficients in the most uniform way.
For this reason, the solutions of \eqref{eq:antisparse_coding} are sometimes referred to as \textit{``antisparse''} or \textit{``spread''} since, contrary to the Lasso problem, they are known to be dense with many \textit{``saturated''} entries satisfying 
\begin{equation}
  \label{eq:lemma:Kruskal_rank}
  \kvbar{\xsol\element{i}} = \kvvbar{\xsol}_\infty.
\end{equation}
Although less popular than its ``sparsity-promoting'' $\ell_1$ counterpart, the ``spreading'' property of the $\ell_\infty$ norm has proved to be of practical interest in several applicative domains, \textit{e.g.}, to design robust analog-to-digital conversion schemes~\cite{Cvetkovic2003,Calderbank2002}, reduce the peak-to-average power ratio in multi-carrier transmissions~\cite{Farrell2009,Ilic2009,Studer2013jsac}, perform approximate nearest neighbor search~\cite{Jegou2012icassp}, outlier detection~\cite{Vural2017} or robust beamforming~\cite{Jiang2018}.
Besides, when combined with a set of linear equality constraints, minimizing an $\ell_\infty$-norm is referred to as the \emph{minimum-effort control problem} in the optimal-control framework~\cite{Neustadt1962,Cadzow1971}.

From a numerical point of view, although many generic convex optimization tools may be straightforwardly applied to problem \eqref{eq:antisparse_coding}, the design of algorithms specifically dedicated to this kind of problems has also focused less attention than its $\ell_1$ counterpart. 
In this contribution, we make one step in that direction by showing that the main principles underlying ``safe screening'' may be extended to problem~\eqref{eq:antisparse_coding}.
More specifically, we emphasize that \eqref{eq:antisparse_coding} can be reduced to a problem of smaller dimension by identifying ``saturated'' components of the solution $\xsol$ and propose simple tests to identify them.
We refer to the proposed methodology as \emph{``safe squeezing''} since it allows to compact several components of $\xsol$ into one single variable.
We show numerically that the proposed safe squeezing procedure can significantly reduce the computational complexity needed to solve the antisparse problem \eqref{eq:antisparse_coding} to some accuracy.

The rest of this paper is organized as follows. 
\Cref{sec:notations} defines the main notations used in this work.
In \Cref{sec:detect_saturation}, we introduce the proposed safe squeezing test and discuss the problem's dimensionality reduction it allows. 
In \Cref{sec:algorithm} we describe two algorithms to solve the equivalent ``reduced'' optimization problem.
In \Cref{sec:num_exp}, we illustrate the performance of the proposed methodology in numerical experiments.
Concluding remarks are finally made in \Cref{sec:conclusion}.

\section{Notations} \label{sec:notations}

Unless otherwise specified, we will use the following convention throughout the paper. 
The vectors are denoted by lowercase bold letters (\textit{e.g.,} $\bfx$) and matrices by uppercase bold letters (\textit{e.g.,} $\bfA$); $\bfx\element{\idxsat}$ refers to the $\idxsat$th component of $\bfx$ and $\bfa_\idxsat$ to the $\idxsat$th column of $\mOp$. 
Given $\xr\in\kR^q$ and $\sat\in\kR$, the shorthand notations ``$\xr\leq w$'' should be understood as ``$\xr\element{\idxsat}\leq \sat$ $\forall \idxsat$''. 
The notations $\ker(\mOp)$  and $\krank(\mOp)$ will be used to denote the null space and the Kruskal rank \cite[p.~56]{Foucart2013} of $\mOp$, respectively.

Caligraphic letters are used to denote sets (\textit{e.g.,} $\satset$).  
$\cset{\satset}$ stands for the complementary set of $\satset$ and $\card(\satset)$ refers to the  cardinality of $\satset$. 

Given a vector $\bfx\in\kR^n$ and a set of indices $\satset\subseteq\{1,\ldots,n\}$, we let $\bfx_\satset$ be the vector of components of $\bfx$ with indices in $\satset$.
Similary, $\bfA_\satset$ denotes the submatrix of $\mOp$ whose columns have indices in $\satset$.
Finally, $\sigma_{\min}(\bfA)$ and $\sigma_{\max}(\bfA)$ refer to the lowest and highest eigenvalue of $\bfA$, respectively.


\section{Detecting and exploiting saturation}
  \label{sec:detect_saturation}

\subsection{Working hypotheses}\label{sec:workingHypothesis}

In the rest of this paper, we will assume that the following working hypothesis is verified:
\begin{align}\label{eq:workingHypothesis}
0< \lambda<\kvvbar{\ktranspose{\mOp}\vobs}_1. 
\end{align}
The left inequality is natural since letting $\lambda=0$ is tantamount to removing the penalization of problem~\eqref{eq:antisparse_coding}. On the other hand, the right inequality ensures that the solution of problem~\eqref{eq:antisparse_coding} is not the all-zero vector ${\bf0}_\dimx$. More specifically, it can be shown that
\begin{align}\label{eq:consequenceWH}
  \kvvbar{\xsol}_\infty > 0\iff \mbox{$\lambda$ verifies \eqref{eq:workingHypothesis}.}
\end{align}
A proof of this result is available in Appendix~\ref{sec:app:nonzerosol}.
Hypothesis~\eqref{eq:workingHypothesis} implies in particular that $\vobs\notin \mathrm{ker}(\ktranspose{\mOp})$ since the two inequalities cannot be satisfied simultaneously  in the opposite case.
In the sequel, we will always assume that~\eqref{eq:workingHypothesis} holds although not explicitly mentioned in our statements.

We will also assume that \eqref{eq:antisparse_coding} admits a unique minimizer $\xsol$. 
Although our subsequent derivations may be quite easily extended to the general case, this working hypothesis greatly simplifies the exposition of our procedure. We will add different comments all along our presentation to point out how the non-uniqueness of the solutions of~\eqref{eq:antisparse_coding} would modify the quantities and results at stake.

\subsection{Dimensionality reduction via saturation detection}  

In this section, we illustrate how the knowledge of the positions of saturated entries in $\xsol$ can lead to memory and complexity savings in the resolution of~\eqref{eq:antisparse_coding}. 
Let\footnote{If $\xsol$ is not unique, $\satsetsol$ must be understood as the set of indices of the components saturating for \textit{all} the solutions of \eqref{eq:antisparse_coding}.}
\begin{subequations}
  \begin{align}
    \satsetsol_+ \;\triangleq\;& \kset{\idxsat}{
      \xsol\element{\idxsat} = + \kvvbar{\xsol}_\infty
    },
    \\
    \satsetsol_- \;\triangleq\;& \kset{\idxsat}{
      \xsol\element{\idxsat} = - \kvvbar{\xsol}_\infty
    }
    ,
  \end{align}
\end{subequations}
be the sets of positive and negative saturated components of ${\xsol}$, and $\satsetsol=\satsetsol_-\cup\satsetsol_+$. 
Then, for any $\satset = \satset_+\cup \satset_-$ with $\satset_+\subseteq \satsetsol_+$ and $\satset_-\subseteq \satsetsol_-$, problem~\eqref{eq:antisparse_coding} can be equivalently rewritten as
\begin{align}
  \label{eq:squeezed_problem}
  (\satsol,\xrsol) \in \, 
  &
  \underset{(\sat,\xr)\in \kR\times \kR^{\dimx-\card(\satset)}}{\arg\min} \;
  \tfrac{1}{2} \kvvbar{\vobs - \mOp_{\cset{\satset}}\xr - \sqatoms\, \sat}_2^2 + \lambda\sat\nonumber\\
  &\text{ subject to } \xr \leq \sat,\,-\xr \leq \sat 
  \stepcounter{equation}
  \tag{$\theequation-\calP^{\text{\texttt{sq}}}_\lambda$}
\end{align}
where
\begin{equation}\label{eq:def:sqatoms}  
  \sqatoms \triangleq \sum_{\idxsat\in\satset_+} \bfa_\idxsat - \sum_{\idxsat\in\satset_-} \bfa_\idxsat.
\end{equation}
In particular, the following bijection holds between the solutions of~\eqref{eq:antisparse_coding} and~\eqref{eq:squeezed_problem}:
\begin{align}
\xsol_{\cset{\satset}}&=\xrsol \label{eq:equivsolA}\\
\xsol\element{\idxsat} &= \label{eq:equivsolB}
  \begin{cases}
    - \sat^\star &\text{ if } \idxsat\in\satset_- \\
    + \sat^\star &\text{ if } \idxsat\in\satset_+.
  \end{cases}
\end{align}
In other words, any minimizer of \eqref{eq:antisparse_coding} defines a minimizer of \eqref{eq:squeezed_problem}, and vice-versa.

The rationale behind problem~\eqref{eq:squeezed_problem} is as follows.
Since any minimizer of \eqref{eq:antisparse_coding} weighs (up to a sign) the elements in $\satset$  by the same value (namely $\kvvbar{\xsol}_\infty$), the signed summation of the columns of $\mOp_{\satset}$ can be done once for all in advance.
This is the meaning of vector $\sqatoms$ in \eqref{eq:def:sqatoms}.
Variable $\sat$ plays the role of the largest absolute value of the elements of $\coeffv$. 
In particular,  the constraints ``$\xr\leq\sat$'' and ``$-\xr\leq\sat$'' ensure that the absolute value of the coefficients weighting the elements of $\mOp_{\cset{\satset}}$ is no larger than $\sat$.
 
We note that, although defining the same set of solutions, the dimensionality of problems~\eqref{eq:antisparse_coding} and~\eqref{eq:squeezed_problem} can be quite different: whereas the former manipulates $\dimx$-dimensional variables, the latter only involves an optimization space of dimension $\dimx-\card(\satset)+1$. 
As a limit case, when $\satset=\satsetsol$, the size of problem~\eqref{eq:squeezed_problem} drops down to $\dimx-\card(\satsetsol)+1$.
The next lemma shows that the dimensionality reduction can be drastic in this case:\footnote{A related result is stated in \cite[Sec.~4.1]{Fuchs2011asilomar}.}
\begin{lemma}
  \label{lemma:Kruskal_rank}
  If $\krank(\mOp)=\dimobs$ then $\card(\satsetsol)\geq \dimx-\dimobs+1$. 
\end{lemma}
\noindent
\Cref{lemma:Kruskal_rank} is a direct consequence of \Cref{th:ONC} below.
The hypothesis ``$\krank(\mOp)=\dimobs$'' holds as soon as any group of $\dimobs$ columns of $\mOp\in\kR^{\dimobs\times \dimx}$ is linearly independent~\cite[p.~56]{Foucart2013}.
This is, for example, a typical setup in machine learning applications where the columns of matrix $\mOp$ contain randomly generated features.\footnote{More particularly, this is the case with probability $1$ when the columns of $\mOp$ are drawn according to any distribution that admits a density with respect to the Lebesgue measure.}
In this case, we see from \Cref{lemma:Kruskal_rank} that the dimension of problem~\eqref{eq:squeezed_problem} is no larger than $\dimobs$ if $\satset=\satsetsol$.
In the overcomplete setting, when $\dimobs\ll\dimx$, solving~\eqref{eq:squeezed_problem} instead of~\eqref{eq:antisparse_coding} then leads to dramatic dimensionality reduction. 
 
A decrease of the size of the problem leads de facto to a reduction of the storage and computational costs necessary to evaluate the solution. 
Tackling problem~\eqref{eq:squeezed_problem} instead of \eqref{eq:antisparse_coding} may thus be obviously of interest to save computational resources. 
The relevance of~\eqref{eq:squeezed_problem} is however conditioned to the identification of a (large) subset of $\satsetsol$. 
In the rest of this section, we propose a procedure, dubbed ``\textit{safe squeezing}'', to perform efficiently this task. 
The term ``\textit{safe}'' refers to the fact that all the indices identified by our procedure necessarily belong to $\satsetsol$. 
The term ``\textit{squeezing}'' relates to the dimensionality reduction allowed by the identification of (some of) the elements of $\satsetsol$.

\subsection{Safe squeezing test}\label{sec:safeSqueezingTests}

In this section, we provide the main ingredient of our safe squeezing procedure. 
It takes the form of a simple test ensuring that an index $\idxsat$ belongs to $\satsetsol$, and is grounded on the following result: 
\begin{theorem}\label{th:ONC}
Let $\satset\subseteq\satsetsol$ and 
\begin{equation}\label{mainsec:eq:dualproblem}
  \stepcounter{equation}
  \tag{\theequation-$\calD_\lambda^\infty$}
  \vdual^\star = \kargmax_{\vdual\in\dualset_{\satset}} \tfrac{1}{2}\kvvbar{\vobs}_2^2-\tfrac{1}{2}\kvvbar{\vobs - \vdual}_2^2
\end{equation}
where 
\begin{equation}
  \label{eq:sat:def:dual_feasibility2}
  \dualset_\satset \triangleq
  \kset{\vdual }{ \kvvbar{\ktranspose{\mOp}_\satset\vdual}_1 + \ktranspose{\sqatoms}\vdual \leq \lambda}
\end{equation}
and $\sqatoms{}$ is defined as in~\eqref{eq:def:sqatoms}.
Then,
\begin{align}\label{th:eq:mainONC}
\forall i:\,  \kvbar{\ktranspose{\mOpv}_\idxsat\vdual^\star}>0 \Rightarrow \idxsat\in\satsetsol_{\mathrm{sign}(\ktranspose{\mOpv}_\idxsat\vdual^\star)}.
\end{align}
Moreover, if $\krank(\mOp)=\dimobs$, $\ktranspose{\mOpv}_\idxsat\vdual^\star\neq0$ for at least $\dimx-\dimobs+1$ $\mOpv_\idxsat$'s. 
\end{theorem}

\noindent
A proof of this result is available in Appendix~\ref{app:proofsMainResults}. 
Theorem~\ref{th:ONC} provides an appealing way of identifying saturated components in $\xsol$: we have from \eqref{th:eq:mainONC} that $\idxsat\in \satsetsol$ as soon as $\mOpv_\idxsat$ is not orthogonal to some vector $\vdual^\star$ defined in \eqref{mainsec:eq:dualproblem}.
Moreover, if $\krank(\mOp)=\dimobs$, then at least $\dimx-\dimobs+1$ saturated components of $\xsol$ can be identified by verifying that $\kvbar{\ktranspose{\mOpv}_\idxsat\vdual^\star}>0$. 

Unfortunately, in practice, finding a maximizer of \eqref{mainsec:eq:dualproblem} turns out to be as difficult as solving our target optimization problem~\eqref{eq:squeezed_problem}. In particular, if $\satset=\emptyset$, addressing \eqref{mainsec:eq:dualproblem} requires the same order of complexity as solving~\eqref{eq:antisparse_coding}. Hence, the direct use of \eqref{th:eq:mainONC} is of poor interest to save computational resources.  Nevertheless, we emphasize below that relaxed versions of \eqref{th:eq:mainONC} can be devised to identify subsets of $\satsetsol$ with a low computational burden.  

Before giving a detailed description of the proposed method, we make two important remarks about $\vdual^\star$. 
First, we see from \eqref{mainsec:eq:dualproblem} that $\vdual^\star$ is defined as a maximizer of a feasible\footnote{We always have $\vdual={\bf0}_\dimobs\in\dualset_\satset$ since $\lambda>0$ by hypothesis.} and strictly-concave problem with a continuous and coercive cost function.
Hence, $\vdual^\star$ always exists \cite[Propositions~A.8]{Bertsekas99ed2} and is unique \cite[Propositions~B.10]{Bertsekas99ed2}. This justifies the equality sign used in \eqref{mainsec:eq:dualproblem}. In fact, as shown in Appendix~\ref{app:demDualProblem}, \eqref{mainsec:eq:dualproblem} corresponds to the Lagrangian dual of  problem \eqref{eq:squeezed_problem}.

Second, the value of $\vdual^\star$ does not depend on the particular subset $\satset\subseteq\satsetsol$ considered in \eqref{mainsec:eq:dualproblem}.  
This fact is only stated here for the sake of keeping the discussion at a reasonable level of complexity, but is proved in Appendix~\ref{app:proofsMainResults}. 
We note that, although the value of $\vdual^\star$ is independent of $\satset$, there may be some computational advantages in considering $\satset$'s with large numbers of elements. 
In particular, we emphasize in Section~\ref{subsec:sat:safe_regions_design} below that the computational cost of the ``dual scaling'' operation, needed to implement the proposed safe squeezing test, evolves linearly with $\card(\cset{\satset})$.

We are now ready to expose the main building block of our safe squeezing procedure. 
Let $\sreg$ be some subset of $\kR^\dimobs$ such that
\begin{align}\label{eq:defSafeRegion}
\vdual^\star\in\sreg. 
\end{align}
For now, we only assume that some $\sreg$ verifying \eqref{eq:defSafeRegion} is available. 
We will see in the next section how to construct such a region. 
A region satisfying \eqref{eq:defSafeRegion} is usually referred to as ``safe region'' in the screening literature, and the same convention will be used hereafter. 

We note that since $\vdual^\star\in\sreg$, we have
\begin{align}
\begin{array}{rcl}
  \displaystyle{\min_{\vdual\in\sreg} \ktranspose{\mOpv}_\idxsat\vdual} &<& \ktranspose{\mOpv}_\idxsat\vdual^\star \\
  \displaystyle{\max_{\vdual\in\sreg} \ktranspose{\mOpv}_\idxsat\vdual} &>& \ktranspose{\mOpv}_\idxsat\vdual^\star, 
\end{array}
\end{align}
so that 
\begin{align}
\begin{array}{lcl}
\displaystyle{\min_{\vdual\in\sreg} \ktranspose{\mOpv}_\idxsat\vdual>0} &\Rightarrow& \ktranspose{\mOpv}_\idxsat\vdual^\star>0\\
\displaystyle{\max_{\vdual\in\sreg} \ktranspose{\mOpv}_\idxsat\vdual<0} &\Rightarrow& \ktranspose{\mOpv}_\idxsat\vdual^\star<0.
\end{array}
\end{align}
Using~\eqref{th:eq:mainONC}, we thus obtain
\begin{align}\label{eq:sqtestgen}
\begin{array}{lcl}
\displaystyle{\min_{\vdual\in\sreg} \ktranspose{\mOpv}_\idxsat\vdual>0} &\Rightarrow& i\in\satsetsol_+ \\
\displaystyle{\max_{\vdual\in\sreg} \ktranspose{\mOpv}_\idxsat\vdual<0} &\Rightarrow& i\in\satsetsol_-.
\end{array}
\end{align}
Hence, if the maximum and minimum values of $\ktranspose{\mOpv}_\idxsat\vdual$ over $\sreg$ are easy to evaluate, the left-hand side of \eqref{eq:sqtestgen} provides a simple test to determine whether $\idxsat$ belongs to $\satsetsol_+$ or $\satsetsol_-$. 
In particular, considering a safe region $\sreg$ with a spherical geometry, that is 
\begin{equation}
  \label{eq:sat:def_sphere}
  \sreg
  = \calB\kparen{\scen, \srad} 
  \triangleq \kset{\vdual\in\kR^\dimobs}{\kvvbar{\vdual-\scen}_2 \leq \srad},
\end{equation}
for some $\scen\in\kR^\dimobs$ and $\srad>0$, leads to  
\begin{align}
\begin{array}{l}
\displaystyle{\min_{\vdual\in\calB\kparen{\scen, \srad}} \ktranspose{\mOpv}_\idxsat\vdual}  = \ktranspose{\mOpv}_\idxsat \scen - \kvvbar{\mOpv_\idxsat}_2 \srad\\
\displaystyle{\max_{\vdual\in\calB\kparen{\scen, \srad}} \ktranspose{\mOpv}_\idxsat\vdual} = \ktranspose{\mOpv}_\idxsat \scen + \kvvbar{\mOpv_\idxsat}_2 \srad. 
\end{array}
\end{align}
Plugging these expressions into \eqref{eq:sqtestgen} yields the following result:
\begin{theorem}[Safe sphere squeezing test] \label{th:SafeSphereSqueezingTest} 
  If $\vdual^\star\in \calB (\scen,\srad)$, then
  \begin{align} \label{eq:SafeSphereSqueezingTest}
    \kvbar{\ktranspose{\mOpv}_\idxsat \scen} >\srad\,\kvvbar{\mOpv_\idxsat}_2 \Rightarrow \idxsat\in\satsetsol_{\sign(\ktranspose{\mOpv}_\idxsat \scen)}. 
  \end{align}
\end{theorem}
\noindent
Squeezing test~\eqref{eq:SafeSphereSqueezingTest}  provides a very practical way of identifying some elements of $\satsetsol$. In particular, testing whether $\idxsat\in\satsetsol$ only requires to evaluate \textit{one} inner product between $\mOpv_\idxsat$ and the center $\scen$ of a safe sphere. 
We note however that passing \eqref{eq:SafeSphereSqueezingTest} is only a \textit{sufficient} condition for $\idxsat\in\satsetsol$. 
In practice, depending on the choice of the center and the radius of the safe region, some elements of $\satsetsol$ may not be identified by the proposed safe squeezing test. 
As a general rule of thumb, ``small'' safe regions will lead to squeezing tests able to identify more saturated components. We will illustrate this behavior in our numerical results in Section~\ref{sec:num_exp}. We elaborate on the construction of ``good'' safe sphere regions in the next subsection. 
For now, let us just notice that if $\scen=\vdual^\star$ and $\srad=0$, $\sreg = \calB(\scen,\srad)$ is safe and its volume is equal to zero. In this case, the safe squeezing test~\eqref{eq:SafeSphereSqueezingTest} boils down to \eqref{th:eq:mainONC}. In particular, provided that $\krank(\mOp)=\dimobs$, at least $\dimx-\dimobs+1$ saturated components of $\xsol$ can be identified (see last part of Theorem~\ref{th:ONC}).  
 
As a final remark, let us mention that Theorem~\ref{th:SafeSphereSqueezingTest} corresponds to \textit{one} specific particularization of~\eqref{eq:sqtestgen} (\textit{i.e.}, to spherical regions~$\sreg$). 
As with procedures derived for Lasso screening, it is also possible to devise squeezing tests based on safe regions having more refined geometries (\textit{e.g.}, dome~\cite{Xiang2012} or truncated dome~\cite{Xiang2017}).
The extension of our safe squeezing procedure to these geometries is however left for future work and will not be considered hereafter.

\subsection{Construction of safe spheres}
  \label{subsec:sat:safe_regions_design}

A cornerstone of the \emph{safe squeezing} methodology presented in Section~\ref{sec:safeSqueezingTests} is the identification of a ``safe'' region, that is a region $\calS$ verifying~\eqref{eq:defSafeRegion}. 
We elaborate on this problem hereafter: 
we emphasize that several safe spheres derived in the context of safe screening for Lasso can be reused for safe squeezing with minor modifications.
We focus here on the spheres ``ST1'' and "GAP" respectively  proposed in \cite{ghaoui2010safe} and~\cite{fercoq2015}. 

Our reasoning is based on two key observations. 
First, the Lasso dual problem shares the same cost function as problem~\eqref{mainsec:eq:dualproblem} but with a definition of the dual feasible set different from~\eqref{eq:sat:def:dual_feasibility2}, see \textit{e.g.},  \cite[Eq. (2)]{malti2016}. 
Second, the safe spheres proposed in \cite{ghaoui2010safe,fercoq2015} are based on convex optimality conditions and the knowledge of some dual feasible point $\vdual$, but never on the specific definition of the dual feasible set. 
Hence, the center and radius of the ``ST1'' and ``GAP'' spheres proposed in \cite{ghaoui2010safe} and~\cite{fercoq2015} can be reused here with the provisio that the dual variable $\vdual$ appearing in these expressions is feasible for \eqref{mainsec:eq:dualproblem} (that is $\vdual\in\dualset_\satset$). 

For example, the ST1 safe sphere derives from the fact that, by definition of $\vdual^\star$:
\begin{align}
\forall \vdual\in \dualset_{\satset}: \kvvbar{\vobs - \vdual}_2\geq \kvvbar{\vobs - \vdual^\star}_2.
\end{align}
Hence, the sphere $\calB\kparen{\scen, \srad}$ with
\begin{align}\label{eq:SS-ST1}
\begin{array}{cl}
\scen &= \vobs\\ 
\srad &= \kvvbar{\vobs - \vdual}_2\\ 
\end{array}
\end{align}
is safe for problem~\eqref{mainsec:eq:dualproblem}. 
We see that expression of the center and the radius in \eqref{eq:SS-ST1} are the same as those proposed in \cite{ghaoui2010safe} with the difference that $\vdual$ must here be feasible for \eqref{mainsec:eq:dualproblem} rather than for the Lasso dual problem. 

Similarly, the expression of the GAP safe sphere can be extended as follows in the context of safe squeezing: 
for any $\satset\subseteq \satsetsol$ and $(\sat,\xr,\vdual)$ verifying
\begin{align}
  \begin{array}{ccc}
    \xr\leq \sat, & -\xr\leq \sat, & \vdual\in\dualset_\satset,
  \end{array}
\end{align}
the sphere $\calB\kparen{\scen, \srad}$ with 
\begin{align}\label{eq:SS_GAP}
  \begin{array}{cl}
    \scen &= \vdual\\ 
    \srad &= \sqrt{2\,\gap(\sat,\xr,\vdual)}\\ 
  \end{array}
\end{align}
and
\begin{align} \label{eq:def_gap}
\gap(\sat,\xr,\vdual) = \tfrac{1}{2} \|\vobs &- \mOp_{\cset{\satset}}\xr - \sqatoms\, \sat\|_2^2 + \lambda\sat \nonumber\\
          &- (\tfrac{1}{2}\kvvbar{\vobs}_2^2-\tfrac{1}{2}\kvvbar{\vobs - \vdual}_2^2)
\end{align}
is safe for problem~\eqref{mainsec:eq:dualproblem}. The proof of this result is similar to that provided in~\cite{fercoq2015} and is omitted here. 
As for the sphere ST1, the expressions of the center and radius remain the same as for the Lasso problem but the dual (resp. primal) variable(s) coming into play must now be feasible for problem~\eqref{mainsec:eq:dualproblem} (resp.~\eqref{eq:squeezed_problem}).

The only computational difficulty in evaluating \eqref{eq:SS-ST1} and \eqref{eq:SS_GAP} stands in the identification of a dual feasible point. 
We may nevertheless identify such a point by extending the ``dual scaling'' procedure proposed in \cite[Section~3.3]{ghaoui2010safe} to the present setup. 
More precisely, given any $\bfz\in\kR^\dimobs$, we clearly have that
\begin{align}\label{eq:dualscaling}
\vdual = \dualscaling(\bfz)
\end{align}
where
\begin{align}
\dualscaling(\bfz) = \nonumber
\left\{
\begin{array}{ll}
\bfz & \mbox{if $\kvvbar{\ktranspose{\mOp}_{\cset{\satset}}\bfz}_1 + \ktranspose{\sqatoms}\bfz\leq0$},\\
\frac{\lambda}{\kvvbar{\ktranspose{\mOp}_{\cset{\satset}}\bfz}_1 + \ktranspose{\sqatoms}\bfz} \bfz& \mbox{otherwise}.
\end{array}
\right.
\end{align}
is dual feasible. 
We note that the implementation of \eqref{eq:dualscaling} requires the computation of $\card(\cset{\satset})+1$ inner products in~$\kR^\dimobs$.

\subsection{Static versus dynamic squeezing} \label{subsec:static_vs_dynamic}

Similarly to screening methods, the proposed safe squeezing procedure can be used either in a \textit{``static''} or a \textit{``dynamic''} way. 

Static squeezing refers to the case where \eqref{eq:SafeSphereSqueezingTest} is applied once for all (with $\satset=\emptyset$) on the columns of matrix $\mOp$ before the application of a numerical optimization procedure. 
In such a case, the dual feasible point $\vdual$ used to build the safe sphere is commonly defined as follows:
\begin{equation}
  \vdual = \dualscaling(\vobs).
\end{equation}
Static screening may for example be of interest to reduce the problem's dimensionality so that its size fits the computer's memory requirements.

Dynamic squeezing consists in interleaving test~\eqref{eq:SafeSphereSqueezingTest} with the iterations of an optimization procedure.
Here, the goal is to refine the quality of the safe region and increase the cardinality of $\satset$ all along the optimization process. 
The general principle of dynamic squeezing (applied to the resolution of~\eqref{eq:antisparse_coding}) is described in \Cref{alg:dynamic_sqeezing_x}.
At each iteration, the parameters of a safe sphere are evaluated by using a couple a primal-dual feasible points $({\coeffv}^{(t-1)},{\vdual}^{(t-1)})$, see line~\ref{alg:dynamic_sqeezing_x:maj_sphere}. 
The safe sphere is then exploited in the squeezing test~\eqref{eq:SafeSphereSqueezingTest} to identify some set $\satset^{(t-\text{\textonehalf})}\subseteq\satsetsol$, see line~\ref{alg:dynamic_sqeezing_x:squeezing_test}. 
In line~\ref{alg:dynamic_sqeezing_x:update_set}, the indices identified at the current iteration are merged to those previously identified. 
Finally, the primal-dual feasible couple is updated in line~\ref{alg:dynamic_sqeezing_x:maj_x}.
This operation is typically carried out by running a few iterations of a numerical procedure addressing problem~\eqref{eq:squeezed_problem} with {$\satset = \satset^{(t)}$ and initialized with the previous iterate $({\coeffv}^{(t-1)},{\vdual}^{(t-1)})$. 
In the common case where the optimization method only returns a primal iterate $\coeffv^{(t)}$, a feasible dual point can for example be computed by dual scaling of $\bfz = \vobs-\mOp\coeffv^{(t)}$.
By construction, we have
\begin{align}
\satset^{(0)}\subseteq \satset^{(1)} \subseteq \ldots \subseteq \satset^{(t)}
\end{align}
so that the dimension of the reduced problem~\eqref{eq:squeezed_problem} considered in line~\ref{alg:dynamic_sqeezing_x:maj_x} does not increase through the iterations of \Cref{alg:dynamic_sqeezing_x}. 
In particular, if $(\coeffv^{(t)},\vdual^{(t)})$ converges to $(\xsol,\vdual^\star)$, the parameters of the GAP safe sphere converge to $(\scen =\vdual^\star ,\srad = 0)$ and, from Theorem~\ref{th:ONC}, $\card(\satset^{(t)}) \geq \dimx - \dimobs+1$ after a sufficient number of iterations when $ \krank(\mOp)= \dimobs$.

\begin{algorithm}[t]
  \DontPrintSemicolon
  \caption[A1]{
  \label{alg:dynamic_sqeezing_x}
  Principle of dynamic squeezing
  }
  \tcp{Initialization}
  ${\bfx}^{(0)}   = {\bf0}_\dimx$, $\satset^{(0)}=\emptyset$ \;
  $\vdual^{(0)} =  \dualscaling(\vobs)$\;
  $t=1$ \tcp{Iteration index}
  \Repeat{convergence criterion is met}{
    \tcp{Squeezing test}\vspace{0.1cm}
    $(\scen^{(t)},\srad^{(t)}) =  \mathrm{sphere\_param}({\coeffv}^{(t-1)},{\vdual}^{(t-1)},\satset^{(t-1)})$ \label{alg:dynamic_sqeezing_x:maj_sphere} \;
    $\satset^{(t-\text{\textonehalf})} = \mathrm{squeezing\_test}(\scen^{(t)},\srad^{(t)})$ \label{alg:dynamic_sqeezing_x:squeezing_test} \;
    $\satset^{(t)} = \satset^{(t-\text{\textonehalf})}\cup \satset^{(t-1)}$ \label{alg:dynamic_sqeezing_x:update_set} \;\vspace{0.2cm}
    \tcp{Iterations of the optimization procedure}\vspace{0.1cm}
    (${\coeffv}^{(t)},{\vdual}^{(t)}) =\mathrm{optim\_update}({\coeffv}^{(t-1)},{\vdual}^{(t-1)} , \satset^{(t)})$  \vspace{0.2cm} \label{alg:dynamic_sqeezing_x:maj_x}\;
    \tcp{Update iteration index}
    $t = t+1$ \vspace{0.2cm}
  }
  \KwOut{${\coeffv}^{(t)}$, $\satset^{(t)}$}
\end{algorithm}


\section{Algorithmic solutions for~\texorpdfstring{\eqref{eq:squeezed_problem}}{(7)}} \label{sec:algorithm}

A crucial difference between ``screening'' and ``squeezing'' is the nature of the optimization problems obtained after reduction. 
Whereas the reduced problem has the same mathematical form as the original one in the context of screening,  the reduced problem~\eqref{eq:squeezed_problem} obtained after squeezing is structurally different from~\eqref{eq:antisparse_coding}. 
As a consequence, numerical procedures devised to solve~\eqref{eq:antisparse_coding} (\textit{e.g.}, \cite{Studer2013jsac}) cannot be used straightforwardly to address~\eqref{eq:squeezed_problem}.

In this section, we describe two algorithmic solutions for problem~\eqref{eq:squeezed_problem}: in \Cref{sec:algorithm:FW} a numerical procedure based on the Frank-Wolfe algorithm \cite[Section 2.2]{Bertsekas99ed2} is proposed; in \Cref{sec:algorithm:proximal_gradient}, we present a strategy based on a rescaled projected gradient method \cite[Section 2.3]{Bertsekas99ed2}.  
These procedures will be used in the next section to assess computational gain allowed by the proposed safe squeezing procedure.  

\subsection{Frank-Wolfe algorithm}\label{sec:algorithm:FW}

Let us consider the following equivalent formulation of~\eqref{eq:squeezed_problem}:
\begin{align}
  \label{eq:bounded_squeezed_problem}
  \min_{(\sat,\xr)\in \kR\times \kR^{\dimx-\card(\satset)}}
  &\tfrac{1}{2} \kvvbar{\vobs - \mOp_{\cset{\satset}}\xr - \sqatoms\, \sat}_2^2 + \lambda\sat\nonumber\\
  &\text{ subject to } 
  \left\{
  \begin{array}{l}
  \xr \leq \sat,\,-\xr \leq \sat, \\
  \sat\leq \ubsat,
  \end{array}
  \right.
\end{align}
where $\ubsat$ is some constant such that $\satsol\leq\ubsat$. We will discuss the identification of $\ubsat$ at the end of this section. 
For now, let us notice that the additional constraint ``$\sat\leq \ubsat$'' does not change the minimizers of the problem but makes the feasible set compact. 
As consequence, 
the conditional gradient method, also known as Frank-Wolfe algorithm \cite{Frank1956}, can be applied to find a numerical solution of \eqref{eq:bounded_squeezed_problem}.  

The recursions of the Frank-Wolfe algorithm particularized to problem~\eqref{eq:bounded_squeezed_problem} are described in \Cref{alg:frank-wolfe:Ced}.
One iteration of the procedure consists of two main steps: the identification of a feasible descent direction and the update of the current iterate.

\begin{algorithm}[t]
  \DontPrintSemicolon
  \caption[A1]{
  \label{alg:frank-wolfe:Ced}
  Frank-Wolfe algorithm for problem~\eqref{eq:bounded_squeezed_problem}.
  }
  \KwIn{$\vobs$, $\mOp_{\cset{\satset}}$, $\sqatoms$, $\ubsat$, $\sat_0$, $\xr_0$}
  $\sat^{(0)}=\sat_0$, $\xr^{(0)} = \xr_0$ \tcp{Initialization}
  $t=1$ \tcp{Iteration index}
  \Repeat{convergence criterion is met}{
    \tcp{Search of a feasible descent direction}\vspace{0.1cm}

    $\sat^{(t-\text{\textonehalf})} = \ubsat$ \label{line:alg:frank-wolfe:Ced:maj_sat} \;
    
    $\xr^{(t-\text{\textonehalf})}   = \ubsat \sign(\ktranspose{\mOp}_{\cset{\satset}}(\vobs  - \mOp_{\cset{\satset}}\xr^{(t-1)} -\sqatoms\, \sat^{(t-1)}))$ \label{line:alg:frank-wolfe:Ced:maj_xr} \;\vspace{0.2cm}
  
    \tcp{Convex update}\vspace{0.1cm}
    
    $\sat^{(t)}   = \cstep \sat^{(t-1)} + (1-\cstep) \sat^{(t-\text{\textonehalf})}$ \label{line:alg:frank-wolfe:Ced:maj_sat_final}\;
    $\xr^{(t)}  \,= \cstep \xr^{(t-1)} \,+ (1-\cstep) \xr^{(t-\text{\textonehalf})}$ \label{line:alg:frank-wolfe:Ced:maj_xr_final} \; \vspace{0.2cm}

    \tcp{Update iteration index}
    $t = t+1$\vspace{0.2cm}

  }
  \KwOut{$\sat^{(t)}$, $\xr^{(t)}$}
\end{algorithm}

The feasible descent direction is identified as the solution of the following optimization problem: 
\begin{align}\label{eq:algorithms:FW:descentdir}
(\sat^{(t-\text{\textonehalf})} , \xr^{(t-\text{\textonehalf})}) =&   
  \kargmin_{(\sat,\xr)}  
  \begin{pmatrix}
  \sat &
  \ktranspose{\xr}
  \end{pmatrix}
  \gradf^{(t-1)}\\
    &\text{ subject to } 
  \left\{
    \begin{array}{l}
      \xr \leq \sat,\,-\xr \leq \sat \\
      \sat\leq \ubsat,
    \end{array}
  \right. \nonumber
\end{align}
where $\gradf^{(t-1)}$ is the gradient of the cost function evaluated at the current iterate $(\sat^{(t-1)},\xr^{(t-1)})$. 
Problem~\eqref{eq:algorithms:FW:descentdir} is linear and admits the closed-form solution given in lines \ref{line:alg:frank-wolfe:Ced:maj_sat}-\ref{line:alg:frank-wolfe:Ced:maj_xr} of~\Cref{alg:frank-wolfe:Ced}. 

The update of the current iterate is performed via a convex combination of $(\sat^{(t-\text{\textonehalf})}, \xr^{(t-\text{\textonehalf})})$ and $(\sat^{(t-1)},\xr^{(t-1)})$, see lines \ref{line:alg:frank-wolfe:Ced:maj_sat_final}-\ref{line:alg:frank-wolfe:Ced:maj_xr_final} in~\Cref{alg:frank-wolfe:Ced}.
Parameter~$\cstep$ is chosen to minimize the cost function over the line segment joining $(\sateq^{(t-\text{\textonehalf})},\xr^{(t-\text{\textonehalf})})$ and $(\sat^{(t-1)}, \xr^{(t-1)})$. 
In the particular setup considered here, this amounts to finding the value of $\cstep$ minimizing a quadratic function over $[0,1]$, which admits a closed-form solution.

We conclude this section by showing that parameter $\ubsat$ can be chosen as follows:
\begin{align}\label{eq:algorithms:FW:defubsat}
  \ubsat = \tfrac{1}{2\lambda} \kvvbar{\vobs - \mOp_{\cset{\satset}}\xr - \sqatoms\, \sat}_2^2 + \sat
\end{align}
where $(\sat,\xr)$ is any feasible point of problem~\eqref{eq:squeezed_problem}.
This follows from the following observation: 
 \begin{align}
  \lambda \satsol 
  &\leq \tfrac{1}{2} \kvvbar{\vobs - \mOp_{\cset{\satset}}\xrsol - \sqatoms\, \satsol}_2^2 + \lambda\satsol\nonumber\\
  &\leq \tfrac{1}{2} \kvvbar{\vobs - \mOp_{\cset{\satset}}\xr - \sqatoms\, \sat}_2^2 + \lambda\sat,\nonumber
\end{align}
where the last inequality is valid for any couple $(\sat,\xr)$ feasible for~\eqref{eq:squeezed_problem}. 
Hence, the definition of $\ubsat$ in~\eqref{eq:algorithms:FW:defubsat} verifies ``$\satsol\leq\ubsat$'' as required at the beginning of this section. 
In particular, $(\sat,\xr)=(0,{\bf0}_{\card(\cset{\satset})})$ is feasible for~\eqref{eq:squeezed_problem} and leads to $\ubsat = \tfrac{\|\vobs\|_2^{2}}{2\lambda}$. We will consider the latter choice in our simulations in Section~\ref{sec:num_exp}.

\subsection{Rescaled projected gradient method}
  \label{sec:algorithm:proximal_gradient}

We consider a rescaled version of~\eqref{eq:squeezed_problem}:
\begin{align}
  \label{eq:rescaled_squeezed_problem}
  \underset{(\sateq,\xr)\in \kR\times \kR^{\dimx-\card(\satset)}}{\min}&\
  \tfrac{1}{2} \kvvbar{\vobs - \mOp_{\cset{\satset}}\xr - \tfrac{\sqatoms}{\scalf}\, \sateq}_2^2 + \tfrac{\lambda}{\scalf}\sateq\\
  &\text{ subject to } \scalf\xr \leq \sateq,\,-\scalf\xr \leq \sateq \nonumber
\end{align}
where $\scalf>0$. The choice of $\scalf$ will be discussed later on in this section and will be made based on convergence rate arguments. 

Particularizing the gradient projection algorithm described in \cite[Section 2.3]{Bertsekas99ed2} to problem~\eqref{eq:rescaled_squeezed_problem} leads to \Cref{alg:rescaled_proj_gradient}.
Each iteration is divided into three main stages: a ``gradient descent'' step (lines \ref{line:alg:rescaled_proj_gradient:maj_xr}-\ref{line:alg:rescaled_proj_gradient:def_z}), a ``projection'' operation (line~\ref{line:alg:rescaled_proj_gradient:projection}) and a ``convex update'' (line~\ref{line:alg:rescaled_proj_gradient:Ced:maj_sat_final}-\ref{line:alg:rescaled_proj_gradient:Ced:maj_xr_final}). 

Starting from a current estimate $(\xr^{(t-1)}, \sateq^{(t-1)})$, the gradient step consists in updating the current iterate in the direction of the negative gradient of the cost function (by an amount $\gstep>0$). 
The step $\eta$ is chosen to minimize the cost function along the direction of the (negative) gradient.
We note that since the cost function is quadratic, the value of $\eta$ admits a simple closed-form expression.

The projection step consists in solving the optimization problem specified in line~\ref{line:alg:rescaled_proj_gradient:projection} of \Cref{alg:rescaled_proj_gradient}. 
Although conceptually simpler than \eqref{eq:rescaled_squeezed_problem}, the latter does not admit any closed-form solution. 
We show nevertheless in Appendix~\ref{sec:app:algoProj} that the procedure described in \Cref{algo:projection_algo} can compute the unique minimizer of this problem in a \textit{finite} number of steps (upper-bounded by $\card(\cset{\satset})$) with a complexity scaling as $\calO(\card(\cset{\satset}) \log(\card(\cset{\satset})))$.
The interested reader is referred to Appendix~\ref{sec:app:algoProj:BoundedNumberIt} for more details on the computational complexity.

Finally, similarly to the Frank-Wolfe algorithm, the update of the current iterate is performed via a convex combination, see lines~\ref{line:alg:rescaled_proj_gradient:Ced:maj_sat_final}-\ref{line:alg:rescaled_proj_gradient:Ced:maj_xr_final} in~\Cref{alg:rescaled_proj_gradient}.
The choice of $\cstep$ is made to minimize the value of the cost function over the line segment joining the two points.

\begin{algorithm}[t]
  \DontPrintSemicolon
  \caption[A1]{
  \label{alg:rescaled_proj_gradient}
  Projected gradient method for problem~\eqref{eq:rescaled_squeezed_problem} 
  }
  \KwIn{$\vobs$, $\mOp_{\cset{\satset}}$, $\sqatoms$, $\scalf$, $\gstep$, $\sateq_0$, $\xr_0$}
  
  $\sateq^{(0)}=\sateq_0$, $\xr^{(0)} = \xr_0$ \tcp{Initialization}
  $t=1$ \tcp{Iteration index}
  
  \Repeat{convergence criterion is met}{
    \tcp{Gradient step}\vspace{0.1cm}
    $\xr^{(t - \twothird)} =  \xr^{(t-1)} + \gstep\, \ktranspose{\mOp}_{\cset{\satset}} \bfz^{(t-1)}$ \label{line:alg:rescaled_proj_gradient:maj_xr} \;
    $\sateq^{(t - \twothird)}  = \sateq^{(t-1)}  + \tfrac{\gstep}{\cst}\kparen{\ktranspose{\sqatoms} \bfz^{(t-1)} - {\lambda}}$ \label{line:alg:rescaled_proj_gradient:maj_sat} \;\vspace{0.1cm}
    where $\bfz^{(t-1)} = \vobs - \mOp_{\cset{\satset}}\xr^{(t-1)} - \tfrac{\sqatoms}{\scalf} \sateq^{(t-1)}$ \; \label{line:alg:rescaled_proj_gradient:def_z}\vspace{0.2cm}

    \tcp{Projection step (solved via \Cref{algo:projection_algo})}\vspace{0.1cm}
    $\displaystyle{(\sateq^{(t - \onethird)},\xr^{(t - \onethird)})}$
    
     $\qquad\ \ = \displaystyle{\kargmin_{\xr,\sateq} \|{\xr - \xr^{(t-\twothird)}}\|_2^2+ (
        \sateq - \sateq^{(t - \twothird)}
    )^2}$
    $\text{\qquad \quad \ subject to } \scalf\xr \leq \sateq,\,-\scalf\xr \leq \sateq \label{line:alg:rescaled_proj_gradient:projection} $\vspace{0.2cm}
    
    \tcp{Convex update} \vspace{0.1cm}
    $\sat^{(t)}   = \cstep \sat^{(t-1)} + (1-\cstep) \sat^{(t-\onethird)}$ \label{line:alg:rescaled_proj_gradient:Ced:maj_sat_final}\;
    $\xr^{(t)}  \,= \cstep \xr^{(t-1)} \,+ (1-\cstep) \xr^{(t-\onethird)}$ \label{line:alg:rescaled_proj_gradient:Ced:maj_xr_final} \;\vspace{0.2cm}

    \tcp{Update iteration index}
    $t = t+1$
  }
  \KwOut{$\xr^{(t)}$, $\sateq^{(t)}$}
\end{algorithm}

Before concluding this section, we elaborate on the choice of the parameters $\scalf$. 
It can be seen (see \cite[Section 2.3.1]{Bertsekas99ed2}), that the speed of convergence of Algorithm~\ref{alg:rescaled_proj_gradient} to the global minimizer of \eqref{eq:rescaled_squeezed_problem} is a direct function of the conditioning of the Hessian of the cost function\footnote{That is the ratio between $\sigma_{\max}(\Hessian)$ and $\sigma_{\min}(\Hessian)$.}
\begin{equation}\label{eq:defHessian}
  \Hessian = 
  \begin{pmatrix}
      \scalf^{-2}\kvvbar{\sqatoms}_2^2 & \kinv{\scalf}\ktranspose{\sqatoms}\mOp_{\cset{\satset}} \\
      \kinv{\scalf}\ktranspose{\mOp}_{\cset{\satset}} \sqatoms & \ktranspose{\mOp}_{\cset{\satset}} \mOp_{\cset{\satset}}
  \end{pmatrix}.
\end{equation}

\noindent
A proper choice of $\scalf$ may thus help  improving the conditioning of this matrix and, consequently, enhance the speed of convergence of the overall procedure. 
In this paper, we suggest the following rule of thumb:
\begin{equation}
  \label{eq:algo:choice_cst:Ced}
  \cst = 
  \begin{cases}
    \kvvbar{\sqatoms}_2 &\text{ if } \sqatoms \neq {\bf0}_m \\
    1 &\text{ otherwise.}
  \end{cases}
\end{equation}
\noindent
This choice is motivated by the ideal case where the columns of $\mOp_{\cset{\satset}}$ and $\sqatoms$ are orthogonal. In this case, the Hessian matrix in~\eqref{eq:defHessian} is diagonal and (because the columns of $\mOp_{\cset{\satset}}$ have unit-norm) we have:
\begin{align}
      \frac{ \sigma_{\min}(\bfH) }{ \sigma_{\max}(\bfH) } = \min\kparen{
        \kinv{\alpha}\kvvbar{\bfs}_2, \alpha\kinv{\kvvbar{\bfs}}_2} . 
\end{align}
Setting $\alpha$ as in~\eqref{eq:algo:choice_cst:Ced} ensures that the conditioning of $\Hessian$ is equal to $1$ in this case.
Although in practice $\mOp_{\cset{\satset}}$ and $\sqatoms$ are commonly not orthogonal, we noticed in our numerical experiments that \eqref{eq:algo:choice_cst:Ced} leads to much better speeds of convergence than the trivial choice $\scalf=1$.

\begin{algorithm}[t]
  \DontPrintSemicolon
  \KwIn{$\xr,\sateq$, $\scalf$} 
  \tcp{Initialization}
  $\sateq^{(0)} = \sateq$\;

  $\calQ^{(0)} \triangleq \{i : \scalf |\xr\element{i}| \geq \sateq^{(0)}\}$\;

  $t=0$ \tcp{Iteration index}\vspace{0.1cm}
  \tcp{Main recursion}

  \Repeat{$\calQ^{(t)}=\calQ^{(t-1)}$}{ \label{line:algo:projection_algo;repeat}
    $t = t + 1$ \label{line:algo:projection_algo;maj_t} \;
    $\displaystyle\sateq^{(t)} = \frac{\scalf{}^2}{\scalf^2 + \card(\calQ^{(t-1)})} \kparen{\sateq + \sum_{ i\in\calQ^{(t-1)}} \frac{\kvbar{\xr\element{i}}}{\scalf} }$\;\label{algo4:line:def wt}
    $\calQ^{(t)} \triangleq \kset{i}{\scalf \kvbar{\xr\element{i}} \geq \sateq^{(t)}}$ \label{line:algo:projection_algo;maj_Q} \;\vspace{0.1cm}
  } \label{line:algo:projection_algo;until} \vspace{0.2cm} 
  \tcp{Final assignment}
  \eIf{$\sateq^{(t)} \leq 0$}{ \label{line:algo:projection_algo;if}
    $\sateq_{\mathrm{proj}} = 0$ and $\xr_{\mathrm{proj}} = {\bf0}$ \label{line:algo:projection_algo;set_sat0}\;
  }{
    $\;\,\sateq_{\mathrm{proj}} \phantom{\element{i}} = \sateq^{(t)}$ \; \label{line:algo:projection_algo;maj_all0}
    $\xr_{\mathrm{proj}}\element{i} =
    \begin{cases}
      \xr\element{i} &\text{if } i\notin\calQ^{(t)}  \\
      \scalf^{-1}\mathrm{sign}(\xr\element{i})\sateq_{\mathrm{proj}} &\text{otherwise} 
    \end{cases}
    $ \label{line:algo:projection_algo;maj_all}\;
  } \label{line:algo:projection_algo;end}
  \KwOut{$\sateq_{\mathrm{proj}},\xr_{\mathrm{proj}}$}
  \caption{
    \label{algo:projection_algo}
    Projection of $(\sateq,\xr)$ onto the feasible set of~\eqref{eq:rescaled_squeezed_problem}.
  }
\end{algorithm}


\section{Numerical experiments}
  \label{sec:num_exp}

In this section, we report several simulation results illustrating the relevance of the proposed squeezing methodology.  
In \Cref{subsec:exp:radius}, we study the effectiveness of the proposed strategy as a function of the ``quality'' of the safe sphere used in its construction.
In \Cref{subsec:exp:complexity,subsec:exp:benchmark}, we demonstrate that safe squeezing can be advocated to accelerate the resolution of problem~\eqref{eq:antisparse_coding}: 
in \Cref{subsec:exp:complexity}, we investigate the computational savings allowed by ``dynamic squeezing'';
in \Cref{subsec:exp:benchmark}, we show that for a given computational budget, safe squeezing enables to reach better convergence properties.

In our simulations, new realizations of the dictionary $\mOp$ and observation vector $\vobs$ are drawn for each trial according to the following distributions. 
The observation vector is generated according to a standard normal distribution. The dictionary obeys one of the following distribution: 
\textit{i}) the entries are i.i.d. realizations of a centered Gaussian; 
\textit{ii}) the entries are i.i.d. realizations of a uniform law on $\kintervcc{0}{1}$; 
\textit{iii}) the rows are randomly-sampled rows of a DCT matrix; \textit{iv}) the rows are shifted versions of a Gaussian curve.
In the following, these four options will be respectively referred to as ``Gaussian'', ``Uniform'', ``DCT'' and ``Toeplitz''.

\subsection{Effectiveness of the safe sphere squeezing test}
  \label{subsec:exp:radius}

{

\newcommand{\highaccsol}{\coeffv_{\texttt{a}}}
\newcommand{\highaccdual}{\vdual_{\texttt{a}}}

\newcommand{\STsphere}{\calB_{\mathrm{ST}1}}
\newcommand{\gapsphere}{\calB_{\mathrm{GAP}}}

In this section, we assess the effectiveness of the proposed squeezing test. 
More specifically, we evaluate the proportion of saturated entries of $\xsol$ which can be identified by the proposed procedure as a function of the ``quality'' of the safe sphere. 

Fig.~\ref{fig:expA:gap_sphere_radius} presents our results.
They have been obtained by repeating the experiment described below $50$ times. 
For each simulation trial, we draw a realization of $\vobs\in\kR^{200}$ and $\mOp\in\kR^{200\times 300}$ according to the distributions described at the beginning of \Cref{sec:num_exp}.
We then compute a ``high-accuracy'' primal-dual solution $(\highaccsol,\highaccdual)$ of \eqref{eq:antisparse_coding}-\eqref{mainsec:eq:dualproblem}.
$\highaccsol$ is evaluated numerically by solving \eqref{eq:antisparse_coding} with \Cref{alg:rescaled_proj_gradient} and a dual gap of $10^{-14}$ as stopping criterion.
$\highaccdual$ is obtained by dual scaling of $\vobs-\mOp\highaccsol$ (see~\eqref{eq:dualscaling}).
This high-accuracy couple is used to construct the following  safe spheres:
\begin{align}
  \bfc_{\mathrm{ST}1} \;=\;& \vobs \label{eq:simu:cST1}\\
  r_{\mathrm{ST}1} \;=\;& r_0 + \kvvbar{\vobs - \highaccdual}_2\label{eq:simu:rST1}
\end{align}
and
\begin{align}
  \bfc_{\mathrm{GAP}} \;=\;& \highaccdual \label{eq:simu:cGAP}\\
  r_{\mathrm{GAP}} \;=\;& r_0 + \sqrt{2\,\gap(\kvvbar{\highaccsol}_\infty,\highaccsol,\highaccdual)}\label{eq:simu:rGAP}
\end{align}
where $\gap$ is defined in~\eqref{eq:def_gap} and $r_0\geq 0$ is some parameter. 
If $r_0=0$, \eqref{eq:simu:cST1}-\eqref{eq:simu:rST1}  and  \eqref{eq:simu:cGAP}-\eqref{eq:simu:rGAP} respectively correspond to the parameters\footnote{Computed with the primal-dual feasible couple $(\highaccsol,\highaccdual)$.} of the ST1 and GAP safe spheres presented in \Cref{subsec:sat:safe_regions_design}. 
By construction, these expressions thus lead to safe spheres for any value of $r_0\geq 0$. 
In our experiments, we consider \eqref{eq:simu:cST1}-\eqref{eq:simu:rGAP} with different $r_0\geq 0$ to simulate different qualities of safe spheres. 
More specifically, if $r_0 = 0$ and $(\highaccsol,\highaccdual)= (\xsol,\vdualsol)$, $r_{\mathrm{ST}1}$ (resp. $r_{\mathrm{GAP}}$) is the smallest possible radius for an ST1 (resp. GAP) safe sphere.
For each value of $r_0$ and each simulation trial, we apply the proposed safe squeezing test~\eqref{eq:SafeSphereSqueezingTest} with the spheres defined in  \eqref{eq:simu:cST1}-\eqref{eq:simu:rGAP}. 
Fig.~\ref{fig:expA:gap_sphere_radius} represents the proportion of saturated entries of $\xsol$ detected by our squeezing test as a function of parameter $r_0$.
The results are averaged over 50 realizations for each value of $r_0$.

\begin{figure}
  \centering
  \begin{subfigure}[b]{.49\columnwidth} 
    \includegraphics[width=\columnwidth]{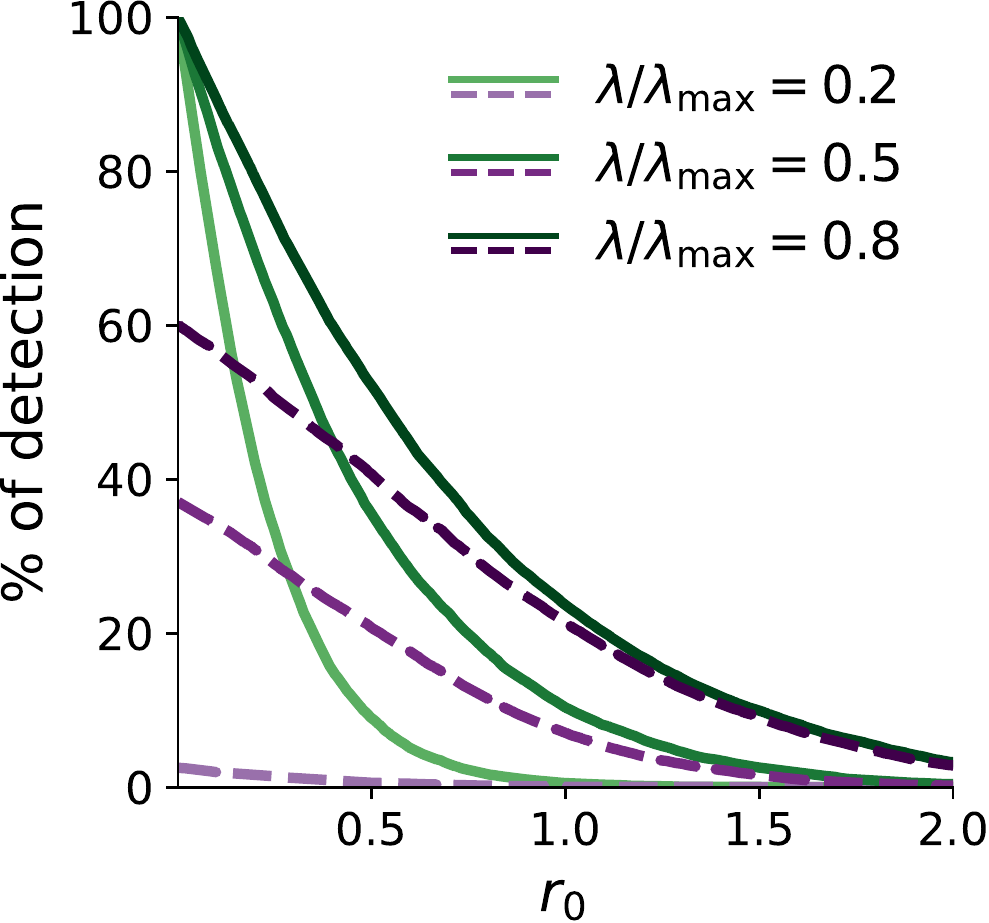}
    \caption{
      \label{fig:expA:gap_sphere_radius:a}
    }
  \end{subfigure}
  \begin{subfigure}[b]{0.49\columnwidth} 
    \includegraphics[width=.9\columnwidth]{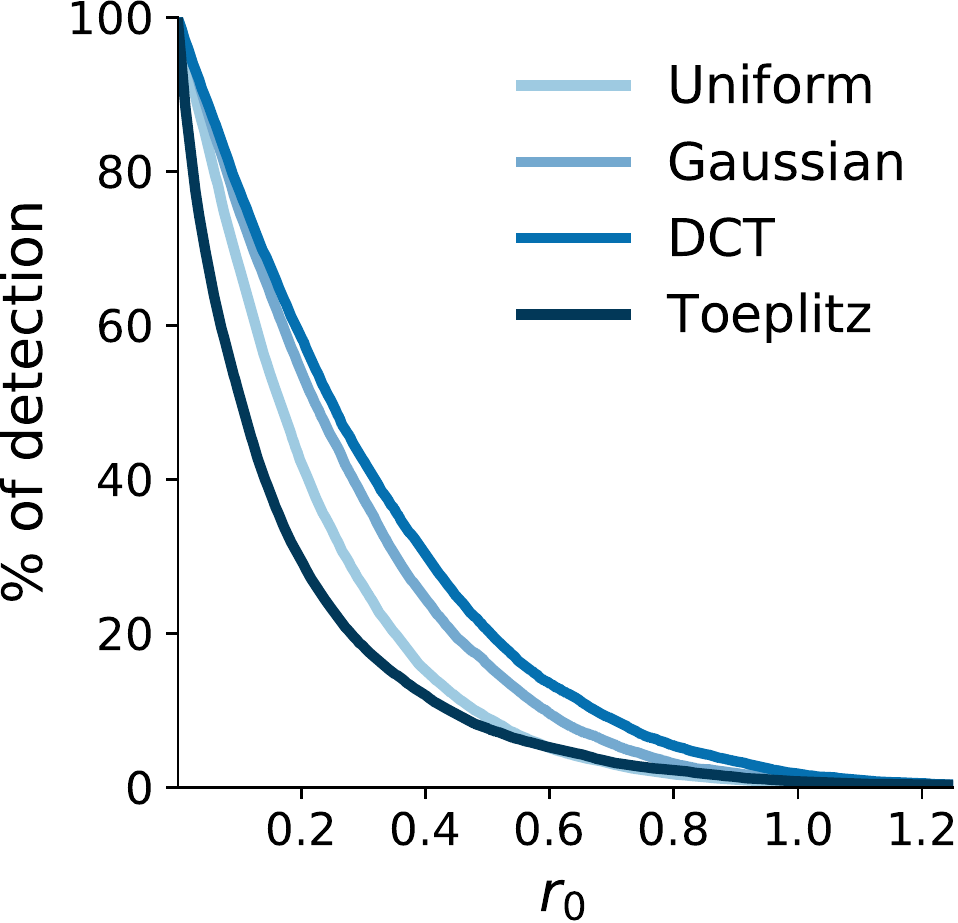}
    \caption{
      \label{fig:expA:gap_sphere_radius:b}
    }
  \end{subfigure}
  \caption{
    \label{fig:expA:gap_sphere_radius}
    Percentage of saturated entries identified with the safe sphere squeezing test as a function of the radius for
    (a) the Gaussian dictionary, $3$ values of $\lambda/ \lambda_{\max}$ using the GAP sphere (thick line) and the ST1 sphere (dashed line)
    (b) the four dictionaries and $\lambda / \lambda_{\max}=0.2$.
  }
\end{figure}

Fig.~\ref{fig:expA:gap_sphere_radius:a} shows the percentage of saturated entries detected with a Gaussian dictionary and three different values of $\lambda/\lambda_{\max}$.
Although we restrict our attention to the Gaussian dictionary, the following comments are valid for the other dictionaries as well.
When $r_0=0$ (\textit{i.e.}, the radius of the safe sphere is equal to $r_{\texttt{GAP}}$ or $r_{\texttt{ST1}}$), one observes that all the saturated entries are detected with the GAP sphere.
However, only a subset of these entries are detected with the ST1 sphere (up to $60\%$ in the best case).
This result was expected since the high accuracy estimate of $\vdualsol$ leads to $\highaccdual \simeq \vdualsol$ and $r_{\mathrm{GAP}}$ is therefore close to zero when $r_0=0$ (see \Cref{subsec:static_vs_dynamic}).
On the other side, the radius of the ST1 sphere is usually bounded away from zero even when $\highaccdual = \vdualsol$ and $r_0=0$.

For the two tests, the number of saturated entries detected decreases as the volume of the safe sphere (parametrized by $r_0$ here) increases.
The performance also seems to depend on $\lambda$: higher values lead to better detection results.

Fig.~\ref{fig:expA:gap_sphere_radius:b} shows the percentage of saturated entries detected for the four dictionaries when $\lambda/\lambda_{\max}=0.2$. 
We observe that some dictionaries seem to be more prone to squeezing.
For example, up to $30\%$ additional saturated entries are detected with the DCT dictionary as compared to the Toeplitz dictionary when $r_0\simeq0.3$.

}

\subsection{Complexity savings}
  \label{subsec:exp:complexity}

{

\newcommand{\indexlbd}{j}
\newcommand{\nblbd}{p}

In this section, we evaluate the overall computational gain induced by the proposed method in the resolution of problem~\eqref{eq:antisparse_coding}.
More precisely, the total number of operations\footnote{We restrict our attention to multiplications since they entail a much higher computational burden than additions in floating-point  arithmetic.} carried out by several algorithms (with and without squeezing) to solve~\eqref{eq:antisparse_coding} to some accuracy are compared. 
In the case where a sequential implementation of the optimization procedures is considered, this figure of merit can be directly related to CPU time needed to solve~\eqref{eq:antisparse_coding}; if parallel implementation is envisaged, the number of operations can be interpreted in terms of ``energy consumption''.

Four numerical procedures are assessed: \textit{(i)} \algoFitra{}, an accelerated proximal gradient algorithm proposed in~\cite{Studer2013jsac}; 
\textit{(ii)} the Frank-Wolfe procedure described in \Cref{alg:frank-wolfe:Ced} with no squeezing ($\satset = \emptyset$); \textit{(iii)} the projected gradient presented in \Cref{alg:rescaled_proj_gradient} with dynamic squeezing (see \Cref{alg:dynamic_sqeezing_x}); \textit{(iv)} the Frank-Wolfe in \Cref{alg:frank-wolfe:Ced} with dynamic screening.  
Procedures \textit{(ii)}, \textit{(iii)} and \textit{(iv)} are respectively denoted as ``\algoFW{}'', ``\algoPGsqueezing{}'', ``\algoFWsqueezing{}'' in the sequel.
Finally, empirical evidence suggests that the GAP sphere test is more effective than ST1, and we concentrate therefore on the former in our experiments.

Since this experiment aims at demonstrating the advantage of resorting to safe squeezing, we only compare procedures belonging to the same family of algorithms and therefore with similar rate of convergence.
In such a setting, the induced reduction in terms of number of operations is most likely only due to safe squeezing.
In particular, we compare \algoFW{} against \algoFWsqueezing{} and \algoFitra{} against \algoPGsqueezing{} since the two couples of procedures belong to the families of Frank-Wolfe and proximal/projection gradient methods, respectively.

The numbers of operations are computed for a decreasing sequence of penalization parameters $\kfamily{\lambda_{\indexlbd}}{\indexlbd=1}^\nblbd$.
Moreover, we consider a ``warm-start'' initialization: for each $\lambda_{\indexlbd}$, the optimization procedure is initialized with the solution obtained for $\lambda_{\indexlbd-1}$.
For $\lambda_1$, all algorithms are initialized with the zero vector $\mathbf{0}\in\kR^n$.
For each run, a stopping criterion in terms of dual gap is used.
To obtain results with the same order of magnitude\footnote{Frank-Wolfe procedures are indeed known to converge much more slowly than proximal/projection gradient methods.}, two values of the dual gap are used: $10^{-7}$ for \algoFitra{} and \algoPGsqueezing{} and $10^{-4}$ for the two Frank-Wolfe procedures (\algoFW{} and \algoFWsqueezing{}).
All results are averaged over $50$ simulation trials.
For each simulation trial, we draw a realization of $\vobs\in\kR^{100}$ and $\mOp\in\kR^{100\times 150}$ according to the distributions described at the beginning of \Cref{sec:num_exp}.

\begin{figure}
  \centering
  \includegraphics[width=\columnwidth]{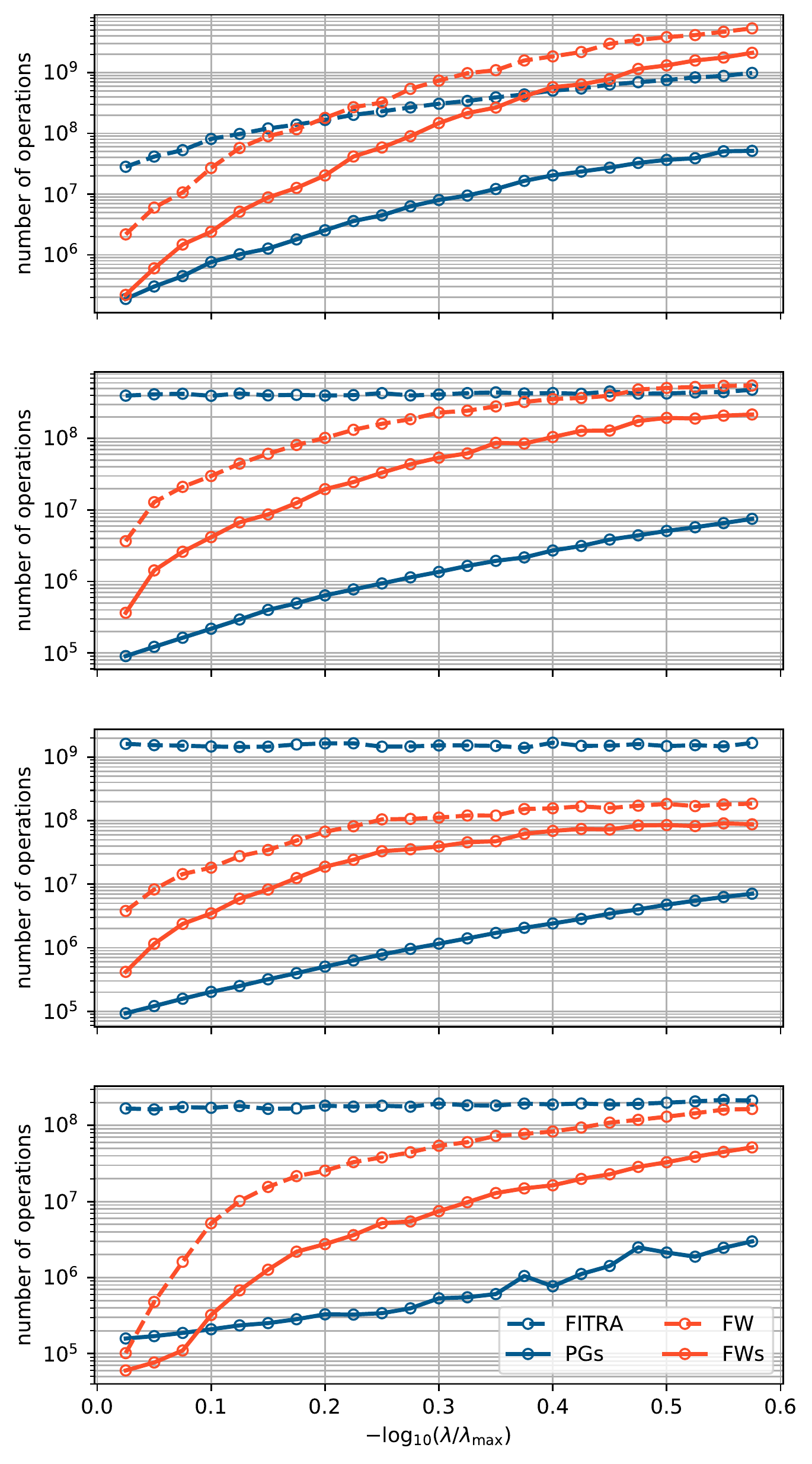}
  \caption{
    \label{fig:computational_gain}
    Number of operations required to achieve convergence for different values of $\lambda/\lambda_{\max}$ and four dictionaries: Uniform (first row), Gaussian (second row), DCT (third row) and Toeplitz (fourth row).
    }
\end{figure}

\begin{figure*}
  \begin{center}
    \includegraphics[width=.95\textwidth]{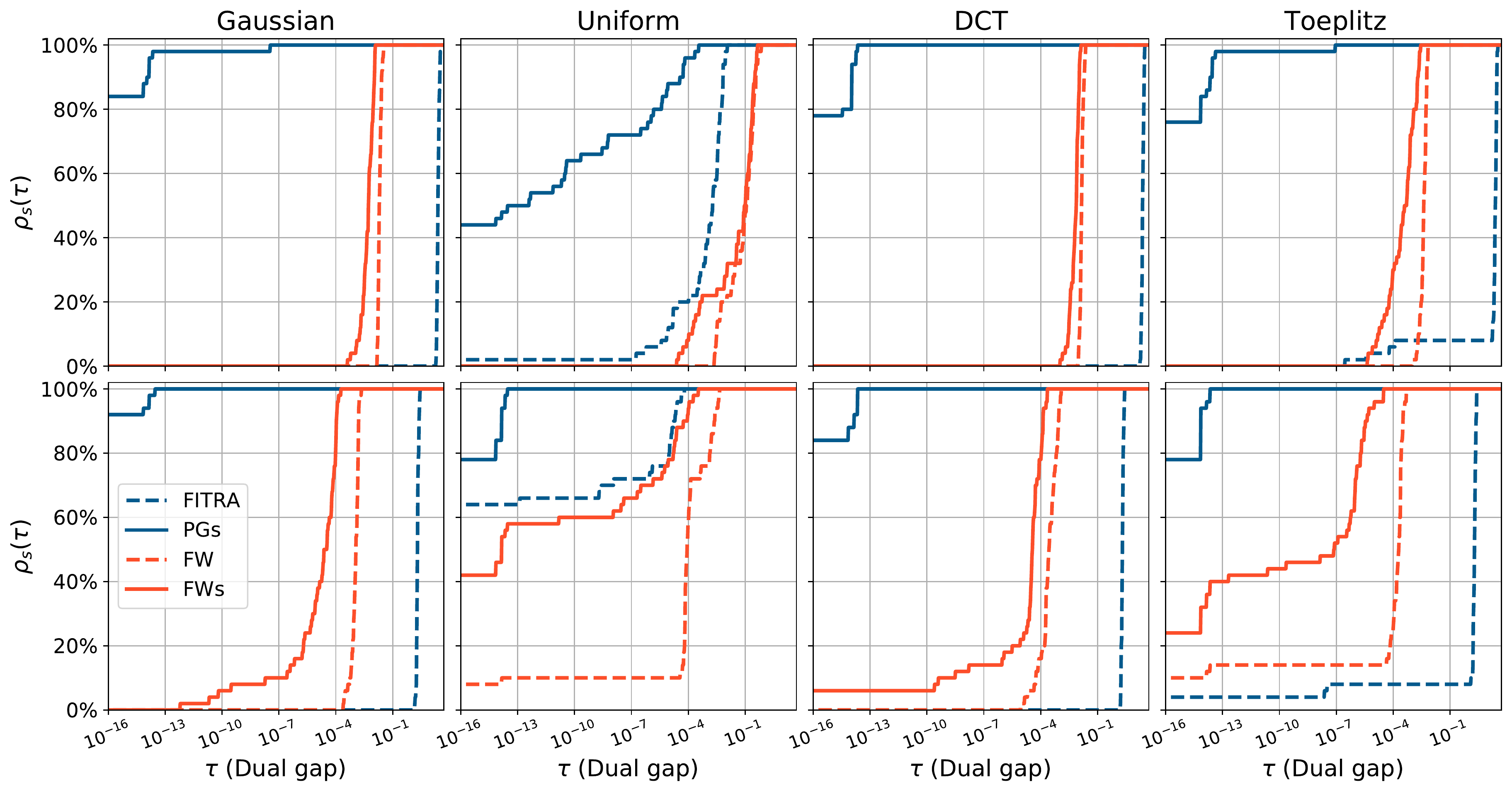}
  \end{center}
  \caption{
    \label{fig:benchmark}
    Performance profiles obtained with the Uniform, Gaussian, DCT and Toeplitz dictionaries. 
    First row: $\displaystyle{{\lambda}/{\lambda_{\max}}=0.3}$; second row: $\displaystyle{{\lambda}/{\lambda_{\max}}=0.8}$.
  }
\end{figure*}

Fig.~\ref{fig:computational_gain} represents the averaged number of operations needed by \algoFitra{}, \algoPGsqueezing{}, \algoFW{} and \algoFWsqueezing{} to achieve the required convergence accuracy.
Each figure corresponds to a different dictionary (Gaussian, Uniform, DCT and Toeplitz).
The curves are plotted as a function of $-\log_{10}(\lambda/\lambda_{\max})$.
We note that, to allow a fair comparison between the different algorithms, we took particular care in the counting of the operations.
In particular, each operation which can be reused latter on in an optimization procedure is not counted twice in our simulation.

In all cases, one observes that \algoPGsqueezing{} (straight blue line) achieves convergence with fewer operations than \algoFitra{} (dashed blue line).
As an example, up to $10^4$ times more operations are required by \algoFitra{} when $-\log_{10}(\lambda/\lambda_{\max})=0.1$ and $\mOp$ is a DCT dictionary (third row).
However, such a ``computational'' saving tends to decrease as $-\log_{10}(\lambda/\lambda_{\max})$ increases.
Such a behavior is in good accordance with typical results of the screening literature where the performance degrades for low values of the penalization parameter.

As far as our simulation setup is concerned, these results show the advantage of considering safe squeezing to reduce the computational burden of computing antisparse representations.
The observed reduction of complexity may be explained by analyzing the cost of one iteration of the four algorithms.
Indeed, since the squeezing tests are done at few additional costs, the implementations of the four considered algorithms results in a cost per iteration of $\calO(mn^{(t)})$ where $n^{(t)} = n-\card(\satset^{(t)})$ and $\satset^{(t)}$ is the number of saturated entries detected at iteration $t$.
While $n^{(t)}$ is decreasing along iterations for \algoPGsqueezing{} and \algoFWsqueezing{}, it remains constant and equal to $n$ for \algoFitra{} and \algoFW{}.
The two mains consequences are \textit{(i)} the computational cost of one iteration is smaller and \textit{(ii)} the induced dimensionality reduction may help improving the conditioning of the optimization problem, resulting potentially in fewer iterations needed to reach convergence.

}

\subsection{Benchmark profiles}
  \label{subsec:exp:benchmark}

{

\newcommand{\nbRun}{p}
\newcommand{\idRun}{j}
\newcommand{\idSolver}{s}

As a final assessment of the performance of the proposed squeezing test, we make use of the Dolan-Moré performance profiles~\cite{Dolan2002}.
For each family of dictionaries, we generate a test set consisting of $\nbRun=50$ random dictionaries $\mOp\in\kR^{100\times150}$ and observations $\vobs\in\kR^{100}$ according to the distributions described at the beginning of \Cref{sec:num_exp}.
Each problem is then solved by using the four algorithms mentioned in the previous section, with a budget of $10^8$ operations.\footnote{The algorithm stops as soon as $10^8$ multiplications have been carried out.}
For each run $\idRun\in\intervint{1}{\nbRun}$ and solvers $\idSolver\in\kbrace{\text{\algoFitra{}, \algoPGsqueezing{}, \algoFW{}, \algoFWsqueezing{}}}$, we define
\begin{equation}
  d_{\idRun, \idSolver} \triangleq \text{dual gap achieved by solver $\idSolver$ at run $\idRun$}.
\end{equation}
In this experiment, we compare the performance profile
\begin{equation}
  \label{eq:exp:def_performance_profile}
  \rho_{\idSolver} (\tau) \triangleq \frac{100}{\nbRun}\card(\kset{\idRun}{d_{\idRun,\idSolver}\leq \tau})
  \quad
  \forall \tau
\end{equation}
of each algorithm $\idSolver$.
In other words, for all possible values of the dual gap $\tau$, $\rho_{\idSolver}(\tau)$ is the (empirical) probability that solver $\idSolver$ reaches a dual gap no greater than $\tau$.

\vspace*{1em}
\Cref{fig:benchmark} presents the performance profiles of the four algorithms with different dictionaries and values of the ratio $\lambda/\lambda_{\max}$. 
As far as our simulation setup is concerned, these results show that for a given budget of operations and any targeted dual gap smaller that $10^{-4}$, resorting to \algoPGsqueezing{} (resp. \algoFWsqueezing{}) allows to solve\footnote{that is, reach a solution with the desired dual gap.} more problems than \algoFitra{} (resp. \algoFW{}).
Moreover, \algoPGsqueezing{} outperforms the other procedures: in all scenarios, at least $45\%$ of the runs reach a dual gap with order of magnitude equal to machine precision ($\tau=10^{-16}$).

As detailed in the last paragraph of \Cref{subsec:exp:complexity}, resorting to safe squeezing may reduce the computational burden of one iteration of an optimization procedure and help improving the conditioning of the optimization problem through an easy rescaling.
With this conclusion in mind, the performance improvements observed in this experiment can be explained by the combination of two following factors.
First, improving the conditioning of the problem results potentially in fewer iterations needed to reach a given dual gap.
Second, reducing the computational cost of one iteration allows to perform more iterations for the same budget of operations, reaching possibly a solution with a lower dual gap.

}


\section{Conclusions}
  \label{sec:conclusion}

In this paper, we proposed a new methodology, dubbed \emph{safe squeezing}, to accelerate algorithms that compute a so-called antisparse representation.
The proposed procedure consists in a set of tests that detect \emph{saturated} entries in the solution of the optimization problem.
These tests rely on the notion of safe region, that is a set that contains the solution of the dual problem.
Benefiting from the recent developments in the screening literature, we showed that two existing safe spheres can be extended in the context of safe squeezing.
The contribution of the saturated entries is then compacted into one single vector, allowing for a potentially dramatic dimensionality reduction.
Computing an antisparse representation becomes equivalent to solving a lower dimensional quadratic problem.
When interleaved with the iterations of an optimization procedure, the approach permits to dynamically detect new saturated entries all along the optimization process.
Numerical simulations showed that the proposed safe squeezing methodology leads to significant computational gains in several numerical setups.


%

\appendices

\section{Proof of the result in~\texorpdfstring{\eqref{eq:consequenceWH}}{(5)}}   \label{sec:app:nonzerosol}

In this section, we prove that~``$\lambda<\kvvbar{\ktranspose{\mOp}\vobs}_1$'' is indeed a necessary and sufficient condition for the solution of~\eqref{eq:antisparse_coding} to be nonzero. 
A direct application of the Fermat’s rule~\cite[Th.~16.2]{Bauschke2011} ensures that $\vcoeff^\star$ is a minimizer of~\eqref{eq:antisparse_coding} if and only if
\begin{equation}
  \label{eq:app:fermat-rule}
  \kinv{\lambda}\ktranspose{\mOp}\kparen{\mOp\vcoeff^\star - \vobs}
  \in\partial\kvvbar{ \vcoeff^\star }_\infty
\end{equation}
where $\partial\kvvbar{ \vcoeff^\star }_\infty$ denotes the sub-differential of the $\ell_\infty$-norm evaluated at $\vcoeff^\star$.
Danskin's theorem~\cite[Th.~B.25]{Bertsekas99ed2} leads to
\begin{equation}
  \label{eq:app-subdifferential-l-infty-norm}
  \partial\kvvbar{ \vcoeff }_\infty
  \triangleq
  \kset{\bfd\in\kR^\dimx}{\kvvbar{\bfd}_1\leq 1 \text{ and } \ktranspose{\vcoeff}\bfd = \kvvbar{\vcoeff}_\infty}
\end{equation}
for all $\vcoeff{}\in\kR^\dimx$.
Result~\eqref{eq:consequenceWH} follows by particularizing~\eqref{eq:app:fermat-rule}-\eqref{eq:app-subdifferential-l-infty-norm} to $\vcoeff^{\star}={\bf0}_\dimx$.

\section{Proof of the results in Section~\ref{sec:detect_saturation}}\label{app:proofsMainResults}
In this section, we provide a proof of Theorem~\ref{th:ONC}.
We also show that the quantity $\vdual^\star$ defined in~\eqref{mainsec:eq:dualproblem} is independent of the set $\satset$. 

In our derivations, we focus on the following generic optimization problem: 
\begin{equation}
  \label{eq:proof:dual:antisparse_screening_aug}
  \min_{
    \substack{
      \bfz\in\kR^{m},\xr\in\kR^{\nsatc} \\
       \sat\geq0
    }
  }
  \tfrac{1}{2}\kvvbar{\vobs - \bfz}_2^2 + \lambda  \sat
  \;\; \text{s.t.}\; 
  \begin{cases}
    \phantom{-}\xr \leq  \sat \\
    -\xr \leq  \sat \\
    \bfz =  \mOps\xr + \sqatoms\sat
    .
  \end{cases}
\end{equation}
Conclusions about problems \eqref{eq:antisparse_coding} and \eqref{eq:squeezed_problem} are then drawn by considering the following substitutions:

\begin{align}
  \mbox{Problem \eqref{eq:antisparse_coding}:}&\ \mOps = \mOp,\ \sqatoms = {\bf0}_m\label{eq:identificationA}\\
  \mbox{Problem \eqref{eq:squeezed_problem}:}&\ \mOps = \mOp_{\cset{\satset}},\ \sqatoms = \sum_{\idxsat\in\satset_+} \bfa_\idxsat - \sum_{\idxsat\in\satset_-} \bfa_\idxsat. \label{eq:identificationB} 
\end{align}
We organize the presentation of the results as follows. 
In Section~\ref{app:demDualProblem} we derive the Lagrangian dual problem of \eqref{eq:proof:dual:antisparse_screening_aug}.  
In Section~\ref{app:connectionDualVariables}, we emphasize some properties of the solutions of the dual problem. 
The latter are then exploited in Section~\ref{sec:app:ONC} to derive (the first part of) the result stated in Theorem~\ref{th:ONC}. 
In Section~\ref{sec:app:unicity}, we elaborate on the uniqueness of the maximizer of the dual problem and its independence with respect to the set $\satset$. 
Finally, in Section~\ref{sec:app:numSatComp} we prove the last part of Theorem~\ref{th:ONC}.

\subsection{Dual problem of~\texorpdfstring{\eqref{eq:proof:dual:antisparse_screening_aug}}{}} \label{app:demDualProblem}

Let $\vdualsign[+]\in\kR^m$ and $\vdualsign[-]\in\kR^m$ (resp. $\vdual\in\kR^n$) denote  the dual variables associated to the inequality (resp. equality) constraint(s) in \eqref{eq:proof:dual:antisparse_screening_aug}. 
The Lagrangian function associated to this problem can then be rewritten as
\begin{align}
  \lagrangian(\xr,\bfz, \sat,\vdual,\vdualsign[+],\vdualsign[-]) =
  & \tfrac{1}{2}\kvvbar{\vobs - \bfz}_2^2 + \ktranspose{\vdual}\bfz\nonumber\\
  & + w\kparen{\lambda- \ktranspose{\sqatoms}\vdual- \ktranspose{\bf1}_n \vdualsign[+]  - \ktranspose{\bf1}_n \vdualsign[-]}\nonumber\\
  & + \ktranspose{\xr}\kparen{-\ktranspose{\mOps}\vdual +\vdualsign[+] - \vdualsign[-]}. \nonumber
\end{align}
Moreover, the corresponding dual function is defined as
\begin{equation*}
  \dual(\vdual,\vdualsign[+],\vdualsign[-]) \triangleq \min_{\xr,\bfz, \sat\geq 0}\lagrangian(\xr,\bfz, \sat,\vdual,\vdualsign[+],\vdualsign[-]).
\end{equation*}
We note that since $\vdualsign[+]\geq 0$ and $\vdualsign[-]\geq 0$, we have $\dual(\vdual,\vdualsign[+],\vdualsign[-])=-\infty$ as soon as the following conditions are not satisfied: 
\begin{align}\label{eq:dualfeasconditions}
  \begin{array}{rl}
  \lambda- \ktranspose{\sqatoms}\vdual
        - \ktranspose{\bf1}_n \vdualsign[+]  - \ktranspose{\bf1}_n \vdualsign[-] \;&\geq\; 0, \\
  -\ktranspose{\mOps}\vdual +\vdualsign[+] - \vdualsign[-] \;&=\; {\bf0}_{\nsatc}. 
  \end{array}
\end{align}
Hence, defining
\begin{equation*}
  \jointdualset \triangleq \kset{(\vdual,\vdualsign[+],\vdualsign[-]) }{ \vdualsign[+]\geq 0, \vdualsign[-]\geq 0 \mbox{ and \eqref{eq:dualfeasconditions} holds}},
\end{equation*}
the dual function $\dual(\vdual,\vdualsign[+],\vdualsign[-])$ can be written as
\begin{align}\nonumber
  \dual(\vdual,\vdualsign[+],\vdualsign[-])
  &=    
  \left\{
  \begin{array}{ll}
    \displaystyle
    \tfrac{1}{2} \kvvbar{\vobs}_2^2
    -\tfrac{1}{2} \kvvbar{\vobs - \vdual}_2^2  &\text{if} \; (\vdual,\vdualsign[+],\vdualsign[-])\in \jointdualset\\
      -\infty & \text{otherwise. } 
    \end{array}
    \right.
\end{align}
The dual problem associated to~\eqref{eq:proof:dual:antisparse_screening_aug} thus reads
\begin{equation}\label{eq:dualproblem}
  \max_{(\vdual,\vdualsign[+],\vdualsign[-])\in \jointdualset} \tfrac{1}{2} \kvvbar{\vobs}_2^2
  -\tfrac{1}{2} \kvvbar{\vobs - \vdual}_2^2. 
\end{equation}

\noindent
From \cite[Proposition 5.2.1]{Bertsekas99ed2},\footnote{More specifically, \eqref{eq:proof:dual:antisparse_screening_aug} is a feasible convex problem with linear constraints and the minimum value of the cost function is upper bounded.} there is no duality gap between~\eqref{eq:proof:dual:antisparse_screening_aug} and~\eqref{eq:dualproblem}; moreover the dual problem \eqref{eq:dualproblem} has (at least) one maximizer.
In fact, we will show in \Cref{sec:app:unicity} below that~\eqref{eq:dualproblem} admits \textit{one unique} maximizer, denoted $(\vdual^\star,\vdualsign[+]^\star,\vdualsign[-]^\star)$ hereafter.

\subsection{Connection between \texorpdfstring{$\vdual^\star$}{},\texorpdfstring{$\vdualsign[+]^\star$}{} and \texorpdfstring{$\vdualsign[-]^\star$}{}}
\label{app:connectionDualVariables}

In this section, we emphasize that $\vdualsign[+]^\star$ and $\vdualsign[-]^\star$ are strongly connected to $\vdual^\star$.
More specifically, we have:
\begin{equation}
  \label{eq:u=f(v)0}
  \begin{split}
    \vdualsign[+]^\star\element{\idxsat} = &\phantom{-} \max(0,\ktranspose{\mOpsv}_\idxsat\vdual^\star) \\
    \vdualsign[-]^\star\element{\idxsat} =  &- \min(0,\ktranspose{\mOpsv}_\idxsat\vdual^\star).
  \end{split}
\end{equation}
This connection can be shown as follows.
First, since there is no duality gap (see \Cref{app:demDualProblem}), any couple of primal-dual optimal variables, say $(\satsol,\xrsol,{\bfz}^\star)$ and $(\vdual^\star,\vdualsign[+]^\star,\vdualsign[-]^\star)$, must verify the well-known Karush-Kuhn-Tucker optimality conditions \cite[Proposition~5.1.5]{Bertsekas99ed2}.
In particular, we have 
\begin{subequations}
    \begin{align}
      \vobs - \vdual^\star \;=\;& \bfz^\star \label{eq:LagrandianOptimality1}\\
      - \ktranspose{\mOps}\vdual^\star + \vdualsign[+]^\star - \vdualsign[-]^\star 
      \;=\;& {\bf0}_p
      \label{eq:dualFeasibiility_b} \\
      \vdualsign^\star
      \;\geq\;& 0
      \label{eq:dualFeasibiility_c}
      \\
       \forall\idxsat: \quad  \vdualsign[+]^\star\element{\idxsat} \kparen{\xrsol\element{\idxsat} - \satsol} \;=\;& 0 \label{eq:slackness1}\\
       \forall\idxsat: \quad   \vdualsign[-]^\star\element{\idxsat} \kparen{\xrsol\element{\idxsat} + \satsol} \;=\;& 0. \label{eq:slackness2}    
    \end{align}
\end{subequations}
More precisely,~\eqref{eq:LagrandianOptimality1} corresponds to first-order optimality condition while~\eqref{eq:dualFeasibiility_b} and~\eqref{eq:dualFeasibiility_c} partially describe the set of dual feasible variables (see~\eqref{eq:dualfeasconditions}).
One also recognizes in~\eqref{eq:slackness1} and~\eqref{eq:slackness2}  the complementary slackness conditions of the problem.

Using these conditions, we have 
\begin{equation}\label{eq:v+v-=0}
  \vdualsign[+]^\star\element{\idxsat} \vdualsign[-]^\star\element{\idxsat} = 0\ \forall \idxsat
\end{equation}
since
\begin{align}
  \vdualsign[+]^\star\element{\idxsat} >0 
  &\stackrel{\eqref{eq:slackness1} }{\Rightarrow} \xrsol\element{\idxsat} = \satsol \stackrel{\eqref{eq:workingHypothesis}-\eqref{eq:consequenceWH}}{>}0\\
  &\stackrel{\eqref{eq:slackness2}}{\Rightarrow} \vdualsign[-]^\star\element{\idxsat}=0, 
\end{align}
and
\begin{align}
  \vdualsign[-]^\star\element{\idxsat} >0 
  &\stackrel{\eqref{eq:slackness2} }{\Rightarrow} \xrsol\element{\idxsat} = -\satsol \stackrel{\eqref{eq:workingHypothesis}-\eqref{eq:consequenceWH}}{<}0\\
  &\stackrel{\eqref{eq:slackness1}}{\Rightarrow} \vdualsign[+]^\star\element{\idxsat}=0. 
\end{align}
Therefore, combining \eqref{eq:dualFeasibiility_b}, \eqref{eq:dualFeasibiility_c} and \eqref{eq:v+v-=0}, we obtain the following implications:
\begin{align}
  \ktranspose{\mOpsv}_\idxsat\vdual^\star =0 \label{eq:bu=0}
  &\Rightarrow
  \left\{
  \begin{array}{l}
  \vdualsign[-]^\star\element{\idxsat}=0\\
   \vdualsign[+]^\star\element{\idxsat}= 0
  \end{array}
  \right.\\
  \ktranspose{\mOpsv}_\idxsat\vdual^\star >0 \label{eq:bu>0}
  &\Rightarrow
  \left\{
  \begin{array}{l}
  \vdualsign[-]^\star\element{\idxsat}=0\\
   \vdualsign[+]^\star\element{\idxsat}= \ktranspose{\mOpsv}_\idxsat\vdual^\star
  \end{array}
  \right.\\
  \ktranspose{\mOpsv}_\idxsat\vdual^\star <0 \label{eq:bu<0}
  &\Rightarrow
  \left\{
  \begin{array}{l}
   \vdualsign[-]^\star\element{\idxsat}= -\ktranspose{\mOpsv}_\idxsat\vdual^\star\\
  \vdualsign[+]^\star\element{\idxsat}=0.
  \end{array}
  \right.
\end{align}
Expression~\eqref{eq:u=f(v)0} corresponds to a compact reformulation of~\eqref{eq:bu=0},~\eqref{eq:bu>0} and~\eqref{eq:bu<0}.

\subsection{Necessary optimality condition}\label{sec:app:ONC}

We now exploit \eqref{eq:u=f(v)0} together with conditions \eqref{eq:slackness1}-\eqref{eq:slackness2} to prove the necessary optimality condition~\eqref{th:eq:mainONC} stated in Theorem~\ref{th:ONC}. Let us first show that $\vdual^\star$ can be expressed as
\begin{equation}\label{eq:dualprobleminter}
  \vdual^\star = 
  \kargmax_{\vdual\in\dualset}
  \tfrac{1}{2} \kvvbar{\vobs}_2^2
    -\tfrac{1}{2} \kvvbar{\vobs - \vdual}_2^2
\end{equation}
where 
\begin{equation}
  \label{eq:app:defdualset}
  \dualset \triangleq
  \kset{\vdual }{ \kvvbar{\ktranspose{\mOps}\vdual}_1 + \ktranspose{\sqatoms}\vdual \leq \lambda}
  .
\end{equation}
To this end, let us define 
\begin{align}
  \calF \triangleq \kset{(\vdual,\vdualsign[+],\vdualsign[-]) }{
  \begin{array}{rr}
  \vdualsign[+]\element{\idxsat} =&  \max(0,\ktranspose{\mOpsv}_\idxsat\vdual)\\
  \vdualsign[-]\element{\idxsat} =&- \min(0,\ktranspose{\mOpsv}_\idxsat\vdual)
  \end{array}
  }.
\end{align}
Clearly, from~\eqref{eq:dualproblem} and~\eqref{eq:u=f(v)0} we have that $(\vdual^\star,\vdualsign[+]^\star,\vdualsign[-]^\star)\in\jointdualset\cap\calF$. 
Hence, 
\begin{equation}\label{eq:restricteddualproblem}
  (\vdual^\star,\vdualsign[+]^\star,\vdualsign[-]^\star) 
  \in \kargmax_{(\vdual,\vdualsign[+],\vdualsign[-])\in \jointdualset\cap\calF}   \tfrac{1}{2} \kvvbar{\vobs}_2^2
    -\tfrac{1}{2} \kvvbar{\vobs - \vdual}_2^2.
\end{equation}
Moreover, combining \eqref{eq:dualfeasconditions} and \eqref{eq:u=f(v)0} leads to
\begin{align}\nonumber
  \jointdualset\cap\calF = \kset{(\vdual,\vdualsign[+],\vdualsign[-]) }{ 
  \begin{array}{rr}
  \lambda \geq& \kvvbar{\ktranspose{\mOps}\vdual}_1 + \ktranspose{\sqatoms}\vdual\\
  \vdualsign[+]\element{\idxsat} =&  \max(0,\ktranspose{\mOpsv}_\idxsat\vdual)\\
  \vdualsign[-]\element{\idxsat} =& - \min(0,\ktranspose{\mOpsv}_\idxsat\vdual)
  \end{array}
  }.
\end{align}
Plugging this definition into~\eqref{eq:restricteddualproblem}, we obtain that $\vdual^\star$ is defined by~\eqref{eq:dualprobleminter}-\eqref{eq:app:defdualset}. 
Moreover, $\vdualsign[+]^\star$ and $\vdualsign[-]^\star$ are univocally specified from $\vdual^\star$ via~\eqref{eq:u=f(v)0}. 

We finally show that the following condition holds between the optimal primal and dual variables:
\begin{equation}\label{app:proof:finalONC}
  (\ktranspose{\mOpsv}_\idxsat\vdual^\star) \kparen{\xrsol\element{\idxsat} - \mathrm{sign}(\ktranspose{\mOpsv}_\idxsat\vdual^\star)\satsol}=0.
\end{equation}
Condition~\eqref{th:eq:mainONC} then corresponds to a particular instantiation of~\eqref{app:proof:finalONC} when $\mOps$ and $\sqatoms$ are defined as in~\eqref{eq:identificationB}. 
Relation~\eqref{app:proof:finalONC} can be obtained as follows. Taking the difference between conditions~\eqref{eq:slackness1}-\eqref{eq:slackness2}, we first have
\begin{equation}
  \xrsol\element{\idxsat}(\vdualsign[+]^\star\element{\idxsat}-\vdualsign[-]^\star\element{\idxsat}) -\satsol (\vdualsign[+]^\star\element{\idxsat}+\vdualsign[-]^\star\element{\idxsat}) =0.
\end{equation}
Now, using \eqref{eq:u=f(v)0} we obtain
\begin{subequations}
\begin{align}
  \vdualsign[+]^\star\element{\idxsat}+\vdualsign[-]^\star\element{\idxsat} =&  \kvbar{\ktranspose{\mOpsv}_\idxsat\vdual^\star} \label{eq:v=f(u)-a} \\
  \vdualsign[+]^\star\element{\idxsat}-\vdualsign[-]^\star\element{\idxsat} = & \ktranspose{\mOpsv}_\idxsat\vdual^\star. \label{eq:v=f(u)-b} 
\end{align}
\end{subequations}
Using~\eqref{eq:v=f(u)-a}-\eqref{eq:v=f(u)-b} and the fact that $\kvbar{\ktranspose{\mOpsv}_\idxsat\vdual^\star} = \kparen{\ktranspose{\mOpsv}_\idxsat\vdual^\star}\mathrm{sign}(\ktranspose{\mOpsv}_\idxsat\vdual^\star)$, we finally end up with the desired result.

\subsection{Uniqueness of \texorpdfstring{$\vdual^\star$}{v*} and independence with respect to \texorpdfstring{$\satset$}{I}}\label{sec:app:unicity}

Problem~\eqref{eq:dualprobleminter} admits a unique minimizer since the cost function is continuous, coercive and strictly concave, see \cite[Propositions~A.8 \& B.10]{Bertsekas99ed2}.
This shows the uniqueness of $\vdual^\star$.
We note that the uniqueness of $(\vdual^\star,\vdualsign[+]^\star,\vdualsign[-]^\star)$ claimed in Section~\ref{app:demDualProblem} then straightforwardly follows from property~\eqref{eq:u=f(v)0}.

Setting $\mOps$ and $\sqatoms$ as in \eqref{eq:identificationB}, we now show the independence of $\vdual^\star$ with respect to the definition of $\satset\subseteq\satsetsol$. 
We note that in this case, problem~\eqref{eq:proof:dual:antisparse_screening_aug} is equivalent to~\eqref{eq:squeezed_problem}.
From optimality condition~\eqref{eq:LagrandianOptimality1}, we have
\begin{equation}
  \vdual^\star = \vobs -\bfz^\star. 
\end{equation}
Now, we have by definition that
\begin{equation}
  \bfz^\star = \mOp_{\cset{\satset}} \xrsol + \sqatoms\satsol. 
\end{equation}
Moreover, because problems~\eqref{eq:antisparse_coding} and~\eqref{eq:squeezed_problem} are equivalent (see Section~\ref{sec:detect_saturation}), we have from~\eqref{eq:equivsolA}-\eqref{eq:equivsolB} that 
\begin{equation}
  \label{eq:valzCste}
  \forall\, \satset\subseteq\satsetsol:  \mOp_{\cset{\satset}} \xrsol + \sqatoms\satsol = \mOp\xsol, 
\end{equation}
where $\xsol$ is the unique (see Section~\ref{sec:workingHypothesis}) minimizer of \eqref{eq:antisparse_coding}. 
Combining the last three equalities, we obtain that $\vdual^\star$ does not depend on the choice of $\satset$ as long as $\satset\subseteq\satsetsol$.\footnote{This property holds true in the more generally setup where $\xsol$ is not unique although the proof is slightly more involved.}

\subsection{Minimum number of saturated components}\label{sec:app:numSatComp}

In this section we prove the last part of Theorem~\ref{th:ONC}. 
Since we showed in Section~\ref{sec:app:unicity} that $\vdual^\star$ is independent of the choice of $\satset\subseteq\satsetsol$, we focus on the case where $\satset=\emptyset$ and $\mOps = \mOp$, $\sqatoms = {\bf0}_m$.

We first note that our working hypothesis~\eqref{eq:workingHypothesis} prevents us from having $\vdual^\star= {\bf0}_\dimobs$. 
Indeed, considering the optimality conditions of problem~\eqref{eq:dualprobleminter}, it can be seen that 
\begin{equation}\label{eq:condition_iff_pb_dual_eq}
  \vobs\notin {\ker(\ktranspose{\mOp})} \Rightarrow \vdual^\star \neq {\bf0}_\dimobs.
\end{equation}
Now, as pointed out in Section~\ref{sec:workingHypothesis}, hypothesis~\eqref{eq:workingHypothesis} implies that $\vobs\notin {\ker(\ktranspose{\mOp})}$ and thus necessarily $\vdual^\star\neq {\bf0}_\dimobs$. 

Since $\vdual^\star\neq {\bf0}_\dimobs$, we observe that ``$\ktranspose{\mOpv}_\idxsat\vdual^\star=0$'' can only occur for at most $\dimobs-1$ ${\mOpv}_\idxsat$'s when $\krank(\mOp)=\dimobs$.
Indeed, $\krank(\mOp)=\dimobs$ implies that any subset of $\dimobs$ columns of $\mOp$ are linearly independent. 
On the other hand, the vector space $\calP \triangleq \{\mOpv\in\kR^\dimobs : \ktranspose{\mOpv}\vdual^\star=0\}$ is a $(\dimobs-1)$-dimensional subspace of $\kR^\dimobs$. 
Hence, if more than $\dimobs-1$ columns of $\mOp$ belong to $\calP$, they must be linearly dependent, which is in contradiction with $\krank(\mOp)=\dimobs$.  
This proves the last part of Theorem~\ref{th:ONC}.


\section{Correctness of Algorithm~\ref{algo:projection_algo}}\label{sec:app:algoProj}

In this appendix, we prove that the procedure described in Algorithm~\ref{algo:projection_algo} solves the following projection problem
\begin{align}\label{eq:projProblem}
  {\min_{\xr',\sat'}} (\sat' - \sateq)^2&+\|{\xr' - \xr}\|_2^2\nonumber\\
            &\text{subject to } \scalf\xr' \leq \sat',\,-\scalf\xr' \leq \sat'
\end{align}
in a \textit{finite} number of steps. 
With a slight abuse of notations,\footnote{In particular, $\xrsol$ should not be confused with the minimizer of problem~\eqref{eq:squeezed_problem}.} we denote hereafter the unique minimizer of this problem as $(\sateq^\star,\xrsol)$.
The dimension of $\xr$ will be denoted $\dimxr$ hereafter. 

The appendix is organized as follows.
In Section~\ref{sec:app:algoProj:BoundedNumberIt}, we prove the convergence of Algorithm~\ref{algo:projection_algo} in a number of steps no greater than $\dimxr$ and discuss its computational complexity.
In Section~\ref{sec:app:algoProj:OptCondProjProblem} we elaborate on necessary and sufficient conditions for $(\sateq^\star,\xrsol)$ to be a minimizer of~\eqref{eq:projProblem}. Finally, we exploit these optimality conditions in Section~\ref{sec:app:algoProj:OptAlgoProj} to show that the output of Algorithm~\ref{algo:projection_algo} corresponds to the unique minimizer of~\eqref{eq:projProblem}.

\subsection{Convergence of Algorithm~\ref{algo:projection_algo} in a finite number of steps}\label{sec:app:algoProj:BoundedNumberIt}

We first note that, by construction, the sequence $\kfamily{\sateq^{(t)}}{}$ is non-decreasing. 
As a matter of fact, by definition of $\calQ^{(t)}$ we must have
\begin{equation}\label{eq:app:algoProj:InvariantRecursion}
  \calQ^{(t)}\subseteq \calQ^{(t-1)}. 
\end{equation}
Moreover, if equality occurs in~\eqref{eq:app:algoProj:InvariantRecursion}, then the stopping criterion of Algorithm~\ref{algo:projection_algo} is satisfied and the procedure terminates. Algorithm~\ref{algo:projection_algo} carries out $t$ iterations provided that
\begin{equation}
  \calQ^{(t)} = \calQ^{(t-1)}\subset\calQ^{(t-2)}\subset\ldots\subset\calQ^{(0)}
\end{equation}
and, in particular, 
\begin{equation}\label{eq:app:algoProj:decreaseCardinality}
\card(\calQ^{(t-1)})<\card(\calQ^{(t-2)})<\ldots<\card(\calQ^{(0)}). 
\end{equation}
Since $\card(\calQ^{(t-1)})$ must be nonnegative and $\card(\calQ^{(0)})\leq \dimxr$, we thus have necessarily from~\eqref{eq:app:algoProj:decreaseCardinality} that $t\leq \dimxr$.
This proves the convergence of Algorithm~\ref{algo:projection_algo} in at most $\dimxr$ steps. 

In practice, \Cref{algo:projection_algo} is implemented as follows.
Vector $\xr$ is first sorted by absolute value and its cumulative sum is precomputed. 
This requires a complexity of $\calO(\dimxr \log \dimxr)$.} 
Then, each of the (at most) $q$ repetitions of lines~\ref{line:algo:projection_algo;maj_t}-\ref{algo4:line:def wt}-\ref{line:algo:projection_algo;maj_Q} scales as $\calO(\log (\card (\calQ^{(t)}))$ since the complexity is concentrated in the identification of the components of $\calQ^{(t)}$ verifying ``$\scalf \kvbar{\xr\element{i}} \geq \sateq^{(t+1)}$''. 
Inasmuch as $\card (\calQ^{(t)})\leq \dimxr$, the overall complexity of the proposed procedure is therefore $\calO(\dimxr \log \dimxr)$.

\subsection{Optimality conditions for \texorpdfstring{\eqref{eq:projProblem}}{(78)}}\label{sec:app:algoProj:OptCondProjProblem}

In this section, we derive necessary and sufficient optimality conditions for problem~\eqref{eq:projProblem}.
These conditions will be exploited in the next section to show the correctness of Algorithm~\ref{algo:projection_algo}.

We first note that 
\begin{equation}\label{appC:signPreservation}
  \sign(\xrsol) = \sign(\xr).
\end{equation}
Indeed, $\xrsol$ is also the minimizer of
\begin{equation}
  \min_{\xr'} \|\xr' - \xr\|_2^2 \quad \text{subject to}\quad \pm \cst \xr' \leq \sateq^\star, 
\end{equation}
which admits the following simple analytical expression:
\begin{equation}
 \forall i: \quad \xrsol(i) = \sign(\xr(i)) \min(|\xr(i)|,\sateq^\star). 
\end{equation}
This shows~\eqref{appC:signPreservation}. 

In the rest of the appendix, we will thus focus, without loss of generality, on the case where
\begin{equation}
  \xr(i)\geq 0\quad\forall i.
\end{equation}
We note that this setup can always be obtained by some simple changes of variables. 
In this case, we have from~\eqref{appC:signPreservation} that $\xrsol\geq {\bf0}_\dimxr$ and problem~\eqref{eq:projProblem} can equivalently be expressed as
\begin{equation}
  \label{eq:pbprox:posNV}
  {\min_{\xr'\geq{\bf0}_\dimxr , \sateq'\geq 0}} \; (\sateq' - \sateq)^2 + \kvvbar{\xr' - \xr}_2^2 \;\; \text{s.t.} \;\; \scalf \xr' \leq \sateq'.
\end{equation}
Since~\eqref{eq:pbprox:posNV} is convex with linear inequality constraints, any minimizer $(\sateq^\star,\xr^\star)$ must satisfy the following set of necessary and sufficient optimality conditions \cite[Propositions 5.2.1 and 5.1.5]{Bertsekas99ed2}:
\begin{equation}
  \scalf\xrsol \leq \sateq^\star
  \label{eq:app:prox:optimality:B}
  ,
\end{equation}
and $\exists{\vdualsign[]^\star\geq {\bf0}_{\dimxr}}$ such that 
\begin{align}
  \quad \forall i:\;\;\; 0 =&\; \vdualsign[]^\star\element{i} \kparen{
    \scalf\xrsol\element{i} - \sateq^\star
  } 
  \label{eq:app:prox:optimality:C}
  ,
  \\
  \label{eq:app:prox:optimality:D}
  (\sateq^\star,\xr^\star)=&\kargmin_{\sateq'\geq0,\xr'\geq {\bf0}_\dimxr} \Bigl\{
  (\sateq' - \sateq)^2 + \kvvbar{\xr' - \xr}_2^2\quad \\
  &\quad\quad\quad\quad\quad\quad+ \sum_{i=1}^\dimxr \vdualsign[]^\star\element{i} \kparen{
    \scalf\xr'\element{i} - \sateq'}\Bigr\}\nonumber.
\end{align}
In the rest of this section, we derive other optimality conditions, equivalent to~\eqref{eq:app:prox:optimality:B}-\eqref{eq:app:prox:optimality:D}, which are more amenable to prove the correctness of Algorithm~\ref{algo:projection_algo}.

We first note that~\eqref{eq:app:prox:optimality:D} is a separable nonnegative least-square problem and its minimizer admits the following closed-form expression:
  \begin{align}
    \sateq^\star \;=\;& 
    \max\Big(0, \sateq + \sum_{i=1}^\dimxr\vdualsign[]^\star\element{i}\Big)
    \label{eq:app:prox:min_sateqB}
    \\
    \xrsol\element{i} \;=\;&
    \max\big( 0, \xr\element{i} - \scalf\vdualsign[]^\star\element{i} \big)
    \label{eq:app:prox:min_coeffvCed}
    .
  \end{align}
We then distinguish between the cases ``$\sateq^\star=0$'' and ``$\sateq^\star>0$''. 

From~\eqref{eq:app:prox:optimality:B}, we see that the case ``$\sateq^\star=0$'' may occur only if $\xrsol = {\bf0}_\dimxr$.
Using~\eqref{eq:app:prox:min_sateqB}-\eqref{eq:app:prox:min_coeffvCed}, this leads to 
\begin{align}\nonumber
(\sateq^\star, \xrsol)=(0,{\bf0}_\dimxr) \iff 
\exists \vdualsign[]^\star\geq {\bf0}_{\dimxr} \mbox{ s.t.}
\begin{cases}
  \displaystyle
  \sateq+ \sum_{i=1}^\dimxr\vdualsign[]^\star\element{i} \leq 0,\\
  \scalf^{-1}\xr(i) \leq \vdualsign[]^\star\element{i},
\end{cases}
\end{align}
or more compactly,
\begin{align}\label{eq:app:NSC:w*=0}
(\sateq^\star, \xrsol)=(0,{\bf0}_\dimxr) \iff \sateq+ \scalf^{-1} \sum_{i=1}^\dimxr\xr(i) &\leq 0. 
\end{align}

We next consider the case ``$\sateq^\star>0$''.
First, this assumption together with~\eqref{eq:app:prox:min_sateqB} leads to
\begin{align}\label{eq:app:w*vsv*}
    \sateq^\star
    \;=\; & \sateq + \sum_{i=1}^\dimxr\vdualsign[]^\star\element{i}. 
\end{align}
Moreover, from~\eqref{eq:app:prox:optimality:C} we have that 
\begin{equation}
  \vdualsign[]^{\star}(i) = 0\quad \forall i\notin\calQ^\star,
\end{equation}
where 
\begin{equation}
  \calQ^\star \triangleq \kset{ i }{ \cst \xr^\star\element{i} = \sateq^\star}. 
\end{equation}
\eqref{eq:app:w*vsv*} thus further simplifies to 
\begin{align}
    \sateq^\star \;=\;& \sateq + \sum_{i\in\calQ^\star}\vdualsign[]^\star\element{i}. 
    \label{eq:app:prox:min_sateqC}
\end{align}
On the other hand, we have (by definition of $\calQ^\star$) that
\begin{equation}\label{eq:app:relxrsolsateqstar}
  \xrsol(i) = \scalf^{-1} \sateq^\star \quad \forall i\in\calQ^\star
\end{equation}
so that, invoking again our hypothesis ``$\sateq^\star>0$'', we obtain from~\eqref{eq:app:prox:min_coeffvCed}
\begin{align}
  \xrsol\element{i} \;=\;&
  \xr\element{i} - \scalf\vdualsign[]^\star\element{i}\quad \forall i\in\calQ^\star. \label{eq:app:relxrsolvdual}
\end{align}
Finally, combining~\eqref{eq:app:prox:min_sateqC},~\eqref{eq:app:relxrsolsateqstar} and~\eqref{eq:app:relxrsolvdual} leads to
\begin{equation}
  \sateq^\star = \frac{\cst^2}{\cst^2 + \card(\calQ^\star)} \kparen{ \sateq + \cst^{-1 }\sum_{i\in\calQ^\star} \xr\element{i} }. 
  \label{eq:prox:optimality_sateqCed}   
\end{equation}
As a consequence $(\sateq^\star,\xrsol)$ with $\sateq^\star>0$ is a solution of~\eqref{eq:pbprox:posNV} if and only if $\exists \vdualsign[]^\star\geq{\bf0}_\dimxr$ such that~\eqref{eq:app:prox:min_sateqC}-\eqref{eq:prox:optimality_sateqCed} are satisfied. 

\subsection{Optimality of \texorpdfstring{$(\sateq_{\mathrm{proj}},\xr_{\mathrm{proj}})$}{(w,q)}}\label{sec:app:algoProj:OptAlgoProj}

We finally prove the correctness of Algorithm~\ref{algo:projection_algo}.  
We assume that the algorithm terminates its main recursion (steps~\ref{line:algo:projection_algo;repeat}-\ref{line:algo:projection_algo;until}) after $t$ iterations (that is $\calQ^{(t)}=\calQ^{(t-1)}$) and show that the assignments defined in steps~\ref{line:algo:projection_algo;if}-\ref{line:algo:projection_algo;end} are such that $(\sateq_{\mathrm{proj}},\xr_{\mathrm{proj}})=(\sateq^\star,\xrsol)$. 

First, if $\sateq^{(t)}\leq 0$, we must have (by definition of $\calQ^{(t)}$) that 
\begin{equation}
  \calQ^{(t)} = \{1,\dotsc, \dimxr\}
\end{equation}
since $\xr\geq {\bf0}_\dimxr$. 
Combining this equality with the definition of $\sateq^{(t)}$ in line~\ref{algo4:line:def wt}
of \Cref{algo:projection_algo} leads to 
\begin{equation}
  \sateq^{(t)}
   = \frac{\scalf^2}{\scalf^2 + \dimxr} \kparen{\sateq+ \scalf^{-1}\sum_{i=1}^\dimxr {\xr\element{i} }}.
\end{equation}
Since $\sateq^{(t)}\leq 0$, the last equality implies
\begin{equation}
  \sateq+ \scalf^{-1}\sum_{i=1}^\dimxr {\xr\element{i} }\leq 0.
\end{equation}
In view of~\eqref{eq:app:NSC:w*=0}, we finally have
\begin{equation}
  \sateq^{(t)}\leq 0 \Rightarrow (\sateq^\star, \xrsol)=(0,{\bf0}_\dimxr).
\end{equation}
This corresponds to the assignment in line~\ref{line:algo:projection_algo;if}-\ref{line:algo:projection_algo;set_sat0} of Algorithm~\ref{algo:projection_algo}. 

If $\sateq^{(t)}>0$, it can easily be verified that the optimality conditions~\eqref{eq:app:prox:min_sateqC}-\eqref{eq:prox:optimality_sateqCed} are verified for
\begin{align}
\xrsol &= \xr_{\mathrm{proj}}\nonumber\\
\sateq^\star &= \sateq_{\mathrm{proj}}\nonumber\\
\vdualsign[]^\star(i) &= 
    \begin{cases}
      0 &\text{ if } \scalf \xr_{\mathrm{proj}}\element{i}<\sateq_{\mathrm{proj}} \\
      {\scalf}^{-2}\kparen{\scalf\xr\element{i} - \sateq_{\mathrm{proj}}} &\text{ otherwise},
    \end{cases} \nonumber       
\end{align}
where $\sateq_{\mathrm{proj}}$ and $\xr_{\mathrm{proj}}$ are defined in lines~\ref{line:algo:projection_algo;maj_all0}-\ref{line:algo:projection_algo;maj_all} of Algorithm~\ref{algo:projection_algo}. 
This concludes the proof.


\ifCLASSOPTIONcaptionsoff
  \newpage
\fi



\small

\begin{thebibliography}{10}
\providecommand{\url}[1]{#1}
\csname url@samestyle\endcsname
\providecommand{\newblock}{\relax}
\providecommand{\bibinfo}[2]{#2}
\providecommand{\BIBentrySTDinterwordspacing}{\spaceskip=0pt\relax}
\providecommand{\BIBentryALTinterwordstretchfactor}{4}
\providecommand{\BIBentryALTinterwordspacing}{\spaceskip=\fontdimen2\font plus
\BIBentryALTinterwordstretchfactor\fontdimen3\font minus
  \fontdimen4\font\relax}
\providecommand{\BIBforeignlanguage}[2]{{%
\expandafter\ifx\csname l@#1\endcsname\relax
\typeout{** WARNING: IEEEtran.bst: No hyphenation pattern has been}%
\typeout{** loaded for the language `#1'. Using the pattern for}%
\typeout{** the default language instead.}%
\else
\language=\csname l@#1\endcsname
\fi
#2}}
\providecommand{\BIBdecl}{\relax}
\BIBdecl

\bibitem{Rockafellar1973}
R.~T. Rockafellar, ``A dual approach to solving nonlinear programming problems
  by unconstrained optimization,'' \emph{Mathematical Programming}, vol.~5,
  no.~1, pp. 354--373, Dec. 1973.

\bibitem{bruck1975}
R.~E. Bruck, ``An iterative solution of a variational inequality for certain
  monotone operators in hilbert space,'' \emph{Bull. Amer. Math. Soc.},
  vol.~81, no.~5, pp. 890--892, Sep. 1975.

\bibitem{Gabay1976ADA}
D.~Gabay and B.~Mercier, ``A dual algorithm for the solution of nonlinear
  variational problems via finite element approximation,'' \emph{Computers \&
  Mathematics with Applications}, vol.~2, no.~1, pp. 17 -- 40, 1976.

\bibitem{ghaoui2010safe}
L.~{El Ghaoui}, V.~Viallon, and T.~Rabbani, ``Safe feature elimination in
  sparse supervised learning,'' EECS Dept., University of California at
  Berkeley, Tech. Rep. UC/EECS-2010-126, Sep. 2010.

\bibitem{Foucart2013}
S.~Foucart and H.~Rauhut, \emph{A Mathematical Introduction to Compressive
  Sensing}.\hskip 1em plus 0.5em minus 0.4em\relax Birkhäuser Basel, 2013.

\bibitem{rosset2004boosting}
S.~Rosset, J.~Zhu, and T.~Hastie, ``Boosting as a regularized path to a maximum
  margin classifier,'' \emph{Journal of Machine Learning Research}, vol.~5, no.
  Aug, pp. 941--973, 2004.

\bibitem{fercoq2015}
O.~Fercoq, A.~Gramfort, and J.~Salmon, ``Mind the duality gap: safer rules for
  the {L}asso,'' in \emph{Proc. Int. Conf. Machine Learning (ICML)}, ser.
  Proceedings of Machine Learning Research, vol.~37, Lille, France, Jul. 2015,
  pp. 333--342.

\bibitem{bonnefoy2015}
A.~Bonnefoy, V.~Emiya, L.~Ralaivola, and R.~Gribonval, ``{Dynamic Screening:
  Accelerating First-Order Algorithms for the Lasso and Group-Lasso},''
  \emph{IEEE Trans. Signal Process.}, vol.~63, no.~19, p.~20, 2015.

\bibitem{malti2016}
A.~{Malti} and C.~{Herzet}, ``Safe screening tests for {L}asso based on firmly
  non-expansiveness,'' in \emph{Proc. IEEE Int. Conf. Acoust., Speech, and
  Signal Proces. (ICASSP)}, Mar. 2016, pp. 4732--4736.

\bibitem{herzet2019}
C.~{Herzet}, C.~{Dorffer}, and A.~{Drémeau}, ``Gather and conquer:
  Region-based strategies to accelerate safe screening tests,'' \emph{IEEE
  Trans. Signal Process.}, vol.~67, no.~12, pp. 3300--3315, Jun. 2019.

\bibitem{Ndiaye2015}
E.~Ndiaye, O.~Fercoq, A.~Gramfort, and J.~Salmon, ``Gap safe screening rules
  for sparse multi-task and multi-class models,'' in \emph{Proc. Int. Conf. on
  Neural Information Processing Systems (NIPS)}.\hskip 1em plus 0.5em minus
  0.4em\relax MIT Press, 2015, pp. 811--819.

\bibitem{Ndiaye2017}
------, ``Gap safe screening rules for sparsity enforcing penalties,'' \emph{J.
  Mach. Learn. Res.}, vol.~18, no.~1, pp. 4671--4703, Jan. 2017.

\bibitem{Rakotomamonjy_Gasso_Salmon19}
G.~G. A.~Rakotomamonjy and J.~Salmon, ``Screening rules for {L}asso with
  non-convex sparse regularizers,'' in \emph{Proc. Int. Conf. Machine Learning
  (ICML)}, 2019.

\bibitem{Cvetkovic2003}
Z.~Cvetkovi\'c, ``Resilience properties of redundant expansions under additive
  noise and quantization,'' \emph{IEEE Trans. Inf. Theory}, vol.~29, no.~3, pp.
  644--656, Mar. 2003.

\bibitem{Calderbank2002}
A.~R. Calderbank and I.~Daubechies, ``The pros and cons of democracy,''
  \emph{IEEE Trans. Inf. Theory}, vol.~48, no.~6, pp. 1721--1725, Jun. 2002.

\bibitem{Farrell2009}
B.~Farrell and P.~Jung, ``A {K}ashin approach to the capacity of the discrete
  amplitude constrained {G}aussian channel,'' in \emph{Proc. Int. Conf.
  Sampling Theory and Applications (SAMPTA)}, Marseille, France, May 2009.

\bibitem{Ilic2009}
J.~{Ilic} and T.~{Strohmer}, ``Papr reductioni in ofdm using kashin's
  representation,'' in \emph{IEEE Workshop on Signal Processing Advances in
  Wireless Communications}, Jun. 2009, pp. 444--448.

\bibitem{Studer2013jsac}
C.~Studer and E.~G. Larsson, ``{PAR}-aware large-scale multi-user {MIMO-OFDM}
  downlink,'' \emph{IEEE J. Sel. Areas Comm.}, vol.~31, no.~2, pp. 303--313,
  Feb. 2013.

\bibitem{Jegou2012icassp}
H.~Jegou, T.~Furon, and J.-J. Fuchs, ``Anti-sparse coding for approximate
  nearest neighbor search,'' in \emph{Proc. IEEE Int. Conf. Acoust., Speech,
  and Signal Proces. (ICASSP)}, 2012, pp. 2029--2032.

\bibitem{Vural2017}
M.~{Vural}, P.~{Jung}, and S.~{Stańczak}, ``A new outlier detection method
  based on anti-sparse representations,'' in \emph{Signal Process. and Comm.
  Appl. Conf. (SIU)}, May 2017, pp. 1--4.

\bibitem{Jiang2018}
X.~{Jiang}, J.~{Chen}, H.~C. {So}, and X.~{Liu}, ``Large-scale robust
  beamforming via $\ell _{\infty }$ -minimization,'' \emph{IEEE Trans. Signal
  Process.}, vol.~66, no.~14, pp. 3824--3837, Jul. 2018.

\bibitem{Neustadt1962}
L.~W. Neustadt, ``Minimum effort control systems,'' \emph{J. SIAM Control},
  vol.~1, no.~1, pp. 16--31, 1962.

\bibitem{Cadzow1971}
J.~A. Cadzow, ``Algorithm for the minimum-effort problem,'' \emph{IEEE Trans.
  on Autom. Control}, vol.~16, no.~1, pp. 60--63, 1971.

\bibitem{Fuchs2011asilomar}
J.-J. Fuchs, ``Spread representations,'' in \emph{asilomar}, 2011.

\bibitem{Bertsekas99ed2}
D.~Bertsekas, \emph{Nonlinear Programming}, 2nd~ed.\hskip 1em plus 0.5em minus
  0.4em\relax Athena Scientific, 1999.

\bibitem{Xiang2012}
Z.~J. {Xiang} and P.~J. {Ramadge}, ``{Fast Lasso screening tests based on
  correlations},'' in \emph{Proc. IEEE Int. Conf. Acoust., Speech, and Signal
  Proces. (ICASSP)}, Mar. 2012, pp. 2137--2140.

\bibitem{Xiang2017}
Z.~J. {Xiang}, Y.~{Wang}, and P.~J. {Ramadge}, ``Screening tests for {L}asso
  problems,'' \emph{IEEE Trans. Patt. Anal. Mach. Intell.}, vol.~39, no.~5, pp.
  1008--1027, May 2017.

\bibitem{Frank1956}
M.~Frank and P.~Wolfe, ``An algorithm for quadratic programming,'' \emph{Naval
  Research Logistics Quarterly}, vol.~3, no. 1‐2, pp. 95--110, 1956.

\bibitem{Dolan2002}
E.~D. Dolan and J.~J. Mor{\'e}, ``Benchmarking optimization software with
  performance profiles,'' \emph{Mathematical Programming}, vol.~91, no.~2, pp.
  201--213, Jan. 2002.

\bibitem{Bauschke2011}
H.~H. Bauschke and P.~L. Combettes, \emph{Subdifferentiability}.\hskip 1em plus
  0.5em minus 0.4em\relax New York, NY: Springer New York, 2011, pp. 223--240.

\end{thebibliography}
\end{document}